# Real Time Control of Tandem-Wing Experimental Platform Using Concerto Reinforcement Learning

Zhang Minghao [*], Yang Xiaojun [†], Wang Zhihe [‡], Wang Liang [§]
*School of Aeronautics, Northwestern Polytechnical University, Xi'an 710072, China;*

*National Key Laboratory of Aircraft Configuration Design, Xi'an 710072, China;*

*National Key Laboratory of Strength and Structural Integrity, Xi'an 710072, China*

*Research & Development Institute of Northwestern Polytechnical University in Shenzhen, Shenzhen, Guangdong, 518057, China.*

This paper introduces the CRL2RT algorithm, an advanced reinforcement learning method aimed at improving the real-time control performance of the Direct-Drive Tandem-Wing Experimental Platform (DDTWEP). Inspired by dragonfly flight, DDTWEP's tandem wing structure causes nonlinear and unsteady aerodynamic interactions, leading to complex load behaviors during pitch, roll, and yaw maneuvers. These complexities challenge stable motion control at high frequencies (2000 Hz). To overcome these issues, we developed the CRL2RT algorithm, which combines classical control elements with reinforcement learning-based controllers using a time-interleaved architecture and a rule-based policy composer. This integration ensures finite-time convergence and single-life adaptability. Experimental results under various conditions, including different flapping frequencies and yaw disturbances, show that CRL2RT achieves a control frequency surpassing 2500 Hz on standard CPUs. Additionally, when integrated with classical controllers like PID, Adaptive PID, and Model Reference Adaptive Control (MRAC), CRL2RT enhances tracking performance by 18.3% to 60.7%. These findings demonstrate CRL2RT's broad applicability and superior performance in complex real-time control scenarios, validating its effectiveness in overcoming existing control strategy limitations and advancing robust, efficient real-time control for biomimetic aerial vehicles.

## I. Nomenclature

$\phi_{exp,i}^n$ = the expected flapping angle position of the $i$-th wing for the next $n$ steps
$A_i$ = the desired flapping amplitude for the $i$-th wing
$f$ = the flapping frequency
$\varphi_i$ = the phase difference of the $i$-th wing
$J_{W,YY}$ = moment of inertia about the $Y$-axis, representing the wing's rotational inertia when rotating around its own $Y$-axis
$J_{W,YZ}$ = product of inertia, representing the coupling of inertia between the $Y$-axis and $Z$-axis, which reflects the asymmetry of the wing and the complexity of the inertia tensor
$J_{W,ZZ}$ = moment of inertia about the $Z$-axis, representing the wing's rotational inertia when rotating around its

---

[*] Ph.D. Candidate, School of Aeronautics.
[†] Professor, School of Aeronautics.
[‡] Ph.D. Candidate, School of Aeronautics.
[§] Ph.D. Candidate, School of Aeronautics.

own $Z$-axis

$T_{y,wing,w,i}$ = the torque around the $Y_{w,i}$ axis experienced by the wing surface, generated based on the aerodynamic force of the single wing

$T_{z,wing,w,i}$ = the torque around the $Z_{w,i}$ axis experienced by the wing surface, generated based on the aerodynamic force of the single wing

$C_{tandem,i}$ = the influence coefficient of the interference from tandem wings on the $i$ − th wing. $T_{VTM,i}$ is the wing membrane constraint torque of the i-th wing

$T_{YAW,i}$ = the additional yaw torque

## II. Introduction

Recent advancements in motor technology and fabrication techniques, have significantly enhanced the performance of hover-capable flapping-wing aircraft, thereby demonstrating greater application flexibility[1-7]. Dragonfly-inspired hover-capable flapping-wing aircraft utilize a unique four-wing independent drive mechanism, enhancing maneuverability[8-11], Consequently, various types of dragonfly-inspired aircraft have been developed in recent years, including those employing mechanical structures to generate the reciprocating motions necessary for lift and asymmetric wing movements for control torques[12-14], as well as direct-drive aircraft utilizing miniature servo motors to simultaneously achieve reciprocating motions for lift and asymmetric wing movements for control torques[8, 15]. Among these, direct-drive biomimetic aircraft, with control architectures and manipulations more akin to conventional robotics[16] and leveraging direct-drive characteristics[17-20] for improved performance, have attracted significant research interest[10, 21, 22]. A typical example is the DDD-1 aircraft, developed by the authors' team and illustrated in Fig.1[9, 10, 22-25]. This platform faces significant challenges due to nonlinear, unsteady aerodynamic interactions resulting from its tandem wings[9, 10, 25]. While sufficient lift is generated to enable vertical motion along a track, achieving stable hovering remains challenging owing to the need for more sophisticated control strategies in the presence of additional aerodynamic interference from closely spaced tandem wings compared to direct-drive dual-wing aircraft. To address this issue and maintain similarity with the DDD-1 while circumventing the limitations that existing experiments cannot directly apply results to airborne biomimetic aircraft[26, 27], the Direct-Drive Tandem-Wing Experiment Platform (DDTWEP), as shown in Fig.2, equipped with a six-component balance, has been developed to explore the pitch, roll, and yaw control strategies of four-wing direct-drive biomimetic aircraft under the nonlinear and unsteady aerodynamic interference of tandem wings.

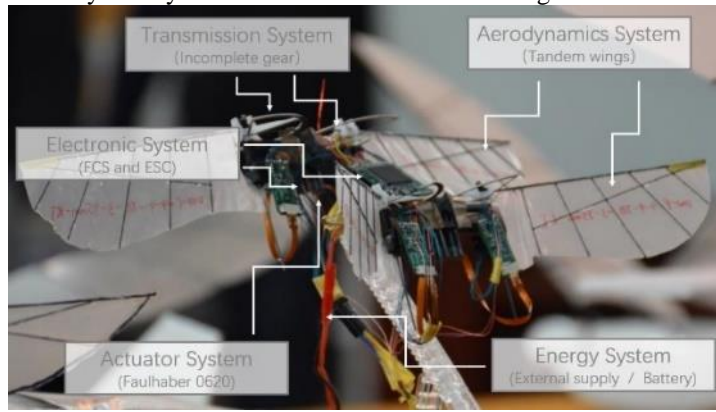

**Fig. 1. Illustration of system integration of the DDD-1**

From the analysis above, a critical challenge for the DDTWEP lies in controlling the mechanical systems under unknown nonlinear and unsteady dynamics in pitch, roll, and yaw operating conditions. Existing research[28-31] frames this problem as a plug-and-play, fully on-the-job, finite-time single-life learning issue, emphasizing safety and efficiency. Additionally, due to the requirement for flapping frequencies to reach 60 Hz to match biological standards[8, 10, 11, 32], the control frequency is set to 2000 Hz[8] to ensure high-precision control, making real-time control issues non-negligible.

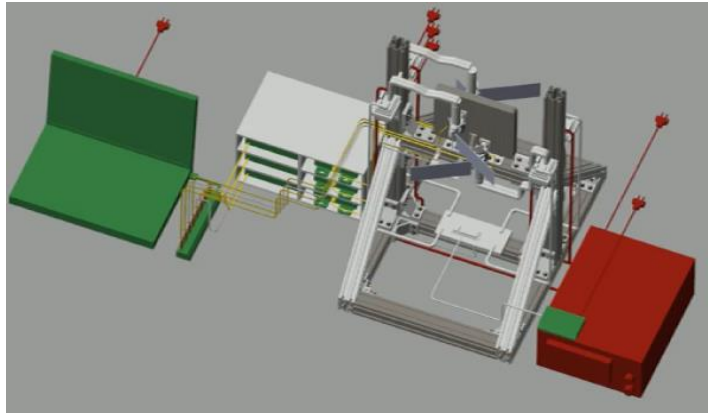

**Fig. 2. Configuration of the DDTWEP**

### A. RELATED WORK
#### 1) Traditional Control Algorithms in Mechanical Systems

Extensive research has focused on controlling mechanical systems using traditional algorithms.

Balakrishnan and Gurfil[33] designed feedback control strategies for low-thrust orbital rendezvous by combining sliding mode and bang-bang control to achieve finite-time stability. Although effective in specific scenarios, this approach relies on simplified assumptions, limiting its adaptability in unpredictable environments.

Jiao et al.[34] proposed an incremental Model Predictive Control (MPC) method to suppress nonlinear pilot-induced oscillations, thereby enhancing flight stability through adaptive control. However, the method's dependence on precise system modeling and high computational demands constrains its real-time adaptability under rapidly changing conditions.

Tsuruta et al.[35] provided theoretical insights into stability in low-thrust orbital contexts by analyzing equilibrium dynamics with the Euler-Lagrange equations. Nonetheless, the reliance on exact mathematical models restricts this method's applicability in environments with unmodeled disturbances.

Beyer et al.[36] enhanced incremental nonlinear dynamic inversion (INLDI) for gust load alleviation in aircraft, reducing structural loads under gust conditions. This approach, however, depends on precise physical models and lacks flexibility in variable environments.

In research directly related to biomimetic aircraft with direct-drive systems, He Ma et al. [8] implemented a 2000 Hz PID controller with manually tuned parameters for dragonfly-inspired aircraft trajectory control, while Xinyan Deng et al.[37] applied Adaptive Robust Control for trajectory tracking in direct-drive aircraft.

Traditional control algorithms, though straightforward, require extensive prior knowledge and data for development, complicating finite-time performance. Moreover, their reliance on precise models reduces robustness against unmodeled disturbances, making them less suitable for online, finite-time, and single-life learning scenarios. This limitation is addressed by the CRL2RT algorithm, which integrates adaptive learning with real-time control, enhancing suitability for dynamic environments.

#### 2) Conventional Reinforcement Learning Control Algorithms in Mechanical Systems

With advancements in edge computing and high-performance training, reinforcement learning (RL) has become increasingly prominent in robotics applications[38, 39]. RL-based controllers, such as Soft Actor-Critic (SAC) [40], Proximal Policy Optimization (PPO) [41], and Twin Delayed Deep Deterministic Policy Gradient (TD3) [42], offer data-driven solutions to the inherent challenges of traditional control methods, promoting improved safety and efficiency[43, 44].

For example, Khai Nguyen et al.[45] proposed an RL-based uncertain switched-variable optimal control framework, focusing on non-autonomous tracking error models to ensure tracking accuracy. This formation control structure, integrating kinematic and dynamic subsystems, has demonstrated stability and optimality through theoretical proofs and simulations. However, the algorithm's dependency on Hamiltonian properties restricts its generalizability. Ziping Wei et al. [46] developed an RL-based trajectory tracking controller for surface vessels, introducing an input saturation penalty function to balance computational cost and performance. Despite these advancements, input saturation constraints introduce specific design challenges.

The following common issues have been identified across these studies:

1. Safety Limitations: None of the algorithms provide adequate safety guarantees against critic estimation errors, which may bias actor updates without effective mitigation mechanisms.

2.Efficiency Limitations: The stability proofs often demonstrate Lyapunov stability but lack finite-time convergence guarantees, resulting in uncertain convergence times.

3.Real-Time Challenges: These algorithms typically struggle to meet stringent real-time motion control demands.

In summary, existing RL-based algorithms require improvements in safety, efficiency, and real-time adaptability, particularly for finite-time single-life control problems. The algorithm proposed in this paper is intended to address these fundamental issues.

### 3) Reinforcement Learning Algorithms for Real time in Mechanical Systems Control

Online real-time performance is a critical requirement in applying reinforcement learning to control typical mechanical systems like the DDTWEP under pitch, roll, and yaw operating conditions[30]. Currently, the primary focus is on average real-time performance, which encompasses the real-time performance of both the inference process and the weight update process, as well as their combined average real-time performance.

Ruturaj Sambhus et al.[47] proposed a real-time, model-free control method based on the PPO, achieving a maximum control frequency of 100 Hz. This method was applied to the closed-loop force control of Series Elastic Actuators under complex 0.1 Hz sine wave and varying frequency chirp force control trajectories, demonstrating performance significantly superior to PID controllers. However, the algorithm's inference real-time performance is limited by its relatively low control frequency, making it unsuitable for inference real-time applications such as sine trajectory tracking at up to 60 Hz in the DDTWEP. Additionally, the reliance on the standard PPO algorithm hinders online continuous weight updates, necessitating system shutdowns for updates.

Shiyu Chen et al.[48] introduced a real-time dynamic response strategy based on deep reinforcement learning, achieving a maximum control frequency of 300 Hz. This approach directly maps drone and gap states to motor thrust, avoiding the cumbersome trajectory planning and independent processing of control modules inherent in traditional optimization methods. A safety-aware exploration mechanism was incorporated, and the method was applied to quadrotor motion control problems, demonstrating strong environmental generalization and adaptability to untrained gap tasks. While the inference real-time performance was enhanced by the direct mapping of states to motor thrust, the algorithm does not support weight updates post-training, limiting the ability to perform fine-tuning or updates based on varying scenarios and preventing online continuous weight updates.

Linfeng Su et al. [49] developed a real-time and optimal hypersonic re-entry guidance method based on Inverse Reinforcement Learning (IRL) to address sparse reward issues by introducing a discriminator network. This method was applied to flight control problems characterized by highly nonlinear dynamics and complex environments. The algorithm achieved an inference speed of 5,988 Hz on high-performance general-purpose computing devices; however, the time required for data transmission was not accounted for. Moreover, like previous methods, weight updates cease post-training, impeding fine-tuning or scenario-based updates and making online continuous weight updates unfeasible.

Kevin Reuer et al. [50] presented a reinforcement learning-based, model-free control method implemented using quantum and FPGA technologies, enabling parallel execution of state observation and inference and achieving a maximum control frequency of 1.304 MHz This method was applied to the control of quantum systems. Although it effectively addresses the inference real-time challenges of reinforcement learning algorithms on dedicated computing devices, its quantum-based implementation is not easily transferable to general-purpose computing devices like CPUs. Additionally, the weight update process is 39,168 times slower than inference, making online continuous weight updates impractical and necessitating system shutdowns for updates.

In summary, while the aforementioned studies achieve high inference speeds up to 7,936 Hz on general-purpose computing devices and 1.304 MHz on dedicated devices—under conditions where weight updates are not performed online, their weight update real-time performance remains inadequate.

### 4) Reinforcement Learning Algorithms for Finite-Time in Mechanical Systems Control

When applying reinforcement learning to control typical mechanical systems such as the DDTWEP under pitch, roll, and yaw operating conditions, finite-time performance is a critical requirement, with a primary focus on training efficiency.

Namhoon Cho et al. [51] presented incremental correction methods for refining neural network parameters or control functions within a continuous-time dynamic system to achieve higher solution accuracy while satisfying intermediate constraints on performance output variables. This approach was applied to the powered descent problem on Mars. However, the algorithm's linearization accuracy remains high only when the system state is close to the reference trajectory. Significant deviations from the reference can render the linearization assumptions invalid, leading to decreased correction accuracy and necessitating additional correction steps to approach the target state. This local linearization method may result in reduced convergence efficiency in highly nonlinear regions, thereby affecting the overall algorithm efficiency and limiting its application potential for motion control of the DDTWEP, especially under numerous random operating conditions.

Osband et al. [52] proposed the Posterior Sampling for Reinforcement Learning (PSRL) algorithm, which operates on repeated events of known duration. At the beginning of each episode, PSRL updates the prior distribution of the Markov decision process and samples from this posterior. The algorithm then follows the optimal policy for the sampled process during execution. Although inspired by posterior sampling rather than optimism, the PSRL algorithm faces challenges in handling potential changes in task characteristics during execution and managing highly nonlinear, complex systems, thereby limiting its applicability to the strongly nonlinear and unsteady load conditions of the DDTWEP.

Jathushan Rajasegaran et al. [29] introduced the Fully Online Meta-Learning (FOML) algorithm, which maintains simplicity by updating only two parameter vectors throughout the online adaptation process: the online parameter vector $\varphi$, updated with each new data batch, and the meta-parameter vector $\theta$, updated correspondingly to accelerate online adaptation and influence online updates through regularization. Despite demonstrating significant multi-task performance, FOML's training is typically conducted in batch settings, making it unsuitable for scenarios with unclear task boundaries or gradually evolving tasks. This limitation restricts its applicability to the DDTWEP under arbitrary pitch, roll, and yaw conditions.

Another approach maximizes using historical experiences to ensure computational efficiency, meeting finite-time requirements. Insights from neuroscience, a key reference source for AI algorithms, demonstrate humans' exceptional adaptability and learning efficiency [53]. Human intelligence exhibits extraordinary abilities to maintain and manipulate memory and skills within activity storage[54, 55]. When learning new tasks, structural reorganization occurs in localized areas responsible for related skills, while other areas remain unchanged[55]. This insight inspired the proposed algorithm, aiming to emulate human rapid learning patterns and apply them to mechanical systems control, especially under the nonlinear and unsteady load conditions of the DDTWEP.

In summary, although the studies have achieved certain improvements and refinements in efficiency, their applicability to the DDTWEP under arbitrary pitch, roll, and yaw conditions remains limited.

5) **Reinforcement Learning Algorithms for Single-Life in Mechanical Systems Control**

Single-life performance is a critical requirement in applying reinforcement learning to control typical mechanical systems such as the DDTWEP under pitch, roll, and yaw operating conditions. This work primarily focuses on robot persistence, the ability to collect data and train continuously with minimal human intervention.

Levine et al. [56] introduced incremental correction methods for refining neural network parameters or control functions within continuous-time dynamic systems to achieve higher solution accuracy while satisfying intermediate constraints on performance output variables. This approach was applied to the powered descent problem on Mars. However, the algorithm's linearization accuracy remains high only when the system state is close to the reference trajectory. Significant deviations from the reference can render the linearization assumptions invalid, leading to decreased correction accuracy and necessitating additional correction steps to approach the target state. This reliance on local linearization may result in reduced convergence efficiency in highly nonlinear regions, thereby affecting overall algorithm efficiency and limiting its application potential for the motion control of the DDTWEP, especially under numerous random operating conditions.

Wallace and Si [57] proposed a decentralized incentivized integral reinforcement learning framework, optimizing a model-based continuous-time reinforcement learning (CT-RL) approach for hypersonic vehicle control. This method incorporates a reference input-driven exploration mechanism and optimizes the pre-scaling of HSV states alongside a decentralized control structure. The new approach demonstrated significantly improved closed-loop stability compared to traditional Linear Quadratic Regulator (LQR) and feedback linearization methods. However, the algorithm relies on the accuracy of system models, making it challenging to handle unmodeled disturbances or dynamic environmental changes. Consequently, this method cannot fully satisfy the single-life control requirements of the DDTWEP under unknown environments or system parameter uncertainties.

Pereira et al. [58] developed a deep learning-based solution that combines nonlinear stochastic optimal control principles with Feynman-Kac theory, utilizing deep forward-backward stochastic differential equations and differentiable neural network layers for the powered descent guidance problem in planetary landings. This approach minimizes fuel consumption and significantly enhances the algorithm's robustness against random disturbances and initial conditions. However, it primarily relies on offline-trained policies suitable for well-defined tasks with known initial conditions. The necessity for extensive pre-training and parameter tuning limits real-time adaptability and self-adjustment in changing environments. Additionally, the algorithm's performance depends on the coverage of initial training data, potentially affecting system robustness and convergence speed under extreme or unforeseen conditions, thus failing to meet the single-life control requirements of the DDTWEP fully.

Yan Yin et al. [59] proposed the Soft Actor-Critic with Curriculum Prioritization and Fuzzy Logic (SCF), enabling mobile robots to navigate efficiently in unpredictable and dynamic environments while ensuring optimal planning control, safety, and robustness. The effectiveness of SCF was validated through experiments in both the Gazebo simulation environment and real-world scenarios. However, the adaptation of appropriate curricula at different learning stages remains challenging. Although trajectory energy was suggested as a sample complexity metric, the effective design and adjustment of these metrics to ensure safety and effectiveness remain problematic. Furthermore, the algorithm does not demonstrate suitability for pitch, roll, and yaw operating conditions. Therefore, achieving safety throughout the process, especially during the initial stages, without introducing complex designs, is a critical functionality addressed in this paper.

Although the aforementioned methods achieve single-life performance through various approaches, they often depend on precise system models or manual designs, failing to satisfy the DDTWEP's single-life control requirements fully.

## B. Related Work and Proposed Algorithm

The adaptive policy learning (APL) framework proposed by Lindsey Kerbel et al. [60] is a foundational reference for this research. APL accelerates learning by utilizing pre-existing, highly engineered default powertrain control (PTC) policies integrated with a dynamically weighted, continuously learning reinforcement learning algorithm, which progressively surpasses the performance of the original policies. This approach introduces a novel RL paradigm that capitalizes on readily available power system data for continuous performance improvement. However, several potential issues have been identified:

1. Lack of Consideration for Online Training and Real-Time Performance: The APL framework employs an updated offline training approach without addressing online updates, resulting in relatively poor real-time performance when applied to new, unknown environments.
2. Limited Direct Interaction with System Dynamics: The baseline algorithm manages most of the system's performance, depriving the RL algorithm of direct interaction with the true system dynamics. This limitation hinders the algorithm's effectiveness in controlling complex dynamics and restricts its applicability to highly complex dynamic learning scenarios.
3. Safety Across Conditions and Lifecycle: Although the dynamic weighting mechanism is innovative, it does not directly address safety issues arising from incorrect actions generated by the RL algorithm and the dynamic weighting mechanism. Early in the RL process, significant action errors may lead to safety issues within the APL framework.

Additionally, the Concerto Reinforcement Learning (CRL) algorithm[21] has been developed to address challenges in safety and efficiency from a finite-time, single-life perspective. CRL introduces two main innovations: a time-interleaved module based on Lipschitz conditions that integrates classical controllers with RL-based controllers to enhance initial stage safety, and a policy composer based on finite-time Lyapunov convergence conditions that organizes past learning experiences to ensure efficiency within finite time constraints. These modules have been validated through ablation experiments. However, CRL primarily targets constant and deterministic conditions, limiting its effectiveness under pitch, roll, and yaw operating conditions. The improved CRL2E algorithm enhances accuracy and convergence speed in random environments but suffers from poor real-time performance, achieving only approximately 600 Hz compared to the desired 2000 Hz.

To address these limitations, particularly the poor real-time performance of the CRL algorithm, the CRL2RT algorithm is proposed. Building upon the CRL framework and adjusting the algorithm architecture from a load-balancing perspective, CRL2RT significantly enhances the inference speed of CRL2E by more than threefold.

The main contributions of this work are outlined as follows:
1. Real-Time Optimization: By leveraging the characteristics of the Concerto Reinforcement Learning architecture and further optimizing the updated neural network weights transmission process, the algorithm framework is enhanced to improve real-time performance.
2. Record-Breaking Control Frequency: To the best of the authors' knowledge, the proposed algorithm achieves the highest control frequency for online weight-updating reinforcement learning control algorithms on general-purpose CPUs, surpassing 2500 Hz while incorporating network weight and data update conditions.
3. Universality of Classical Control Integration: The classical control components of the Concerto Reinforcement Learning algorithm have been tested with three different types of classical controllers: PID, Adaptive PID, and MRAC, further validating the universality of the CRL framework.
4. Extensive Applicability and Performance Enhancement: The CRL2RT algorithm demonstrates broad applicability and performance improvements under various configurations and conditions. In 40 Hz and 60 Hz configurations, the CRL2RT algorithm, combined with PID, Adaptive PID, and MRAC controllers, achieved performance enhancements ranging from 18.3% to 60.7%. In the 20 Hz configuration, significant relative performance improvements were observed when combined with PID and Adaptive PID controllers. Although a convergence trend was noted when combined with the MRAC controller, the performance did not significantly surpass that of the MRAC algorithm. These results collectively validate the broad applicability and potential advantages of the CRL2RT algorithm.

The remainder of this article is structured as follows: Section II describes the problem and introduces the key system models. Section III presents the primary methodologies employed. Section IV introduces the CRL2RT algorithm, detailing its methodology, theoretical underpinnings, and key module compositions. Section V presents the experiments conducted to evaluate the algorithm's performance. Finally, Section VI concludes the article by summarizing the findings and contributions.

## III. Problem Description

### A. Control Problem Description

The problem addressed in this paper is the control of DDTWEP to track various forms of desired trajectories for experimental purposes. To represent the aforementioned desired trajectories (as shown in Equation(1)), the concept of Central Pattern Generators (CPGs)[61] is employed to define the expected motion trajectories of each wing.

$$\phi_{exp,i}^n = A_i \cdot \sin(2\pi f \cdot t + \varphi_i) \tag{1}$$

where $\phi_{exp,i}^n, A_i, f, \varphi_i$ used for this study are provided in the Experimental Settings section.

### B. Modeling direct-drive platform under tandem wing influence

A simulation system has been established to provide the necessary training data for facilitating reinforcement learning. The realism of this simulation system has been validated in existing literature[62]. Based on the model and the electronic prototype form of the experimental platform presented in Fig.2, a simplified framework representing the DDTWEP is constructed as shown in Fig.3, in which DDTWEP is abstracted as a system composed of five rigid bodies with eight degrees of freedom: $\phi_{w,1}, \theta_{w,1}, \phi_{w,2}, \theta_{w,2}, \phi_{w,3}, \theta_{w,3}, \phi_{w,4}, \theta_{w,4}$ and 13 key components: one bench, four motors, four springs, and four wings.

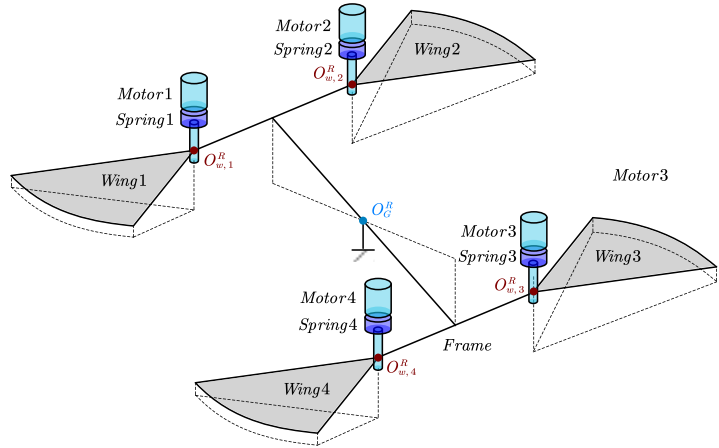

**Fig. 3. Schematic diagram of the main components of the direct-drive tandem-wing experiment platforms**

Following similar existing research[23, 63, 64], the modeling approach and computational system is depicted in Fig.4. The workflow is as follows: First, using the key system models from Table 1 and the Lagrangian equations[28, 65], the system dynamics equations (2) to (15) are constructed. The control algorithm then receives observable variables from the system model and expected commands from the central pattern generators to determine the state. Based on this state, the control algorithm makes decisions and outputs actions. Additionally, rewards are calculated based on the observable variables and expected commands to improve the control algorithm.

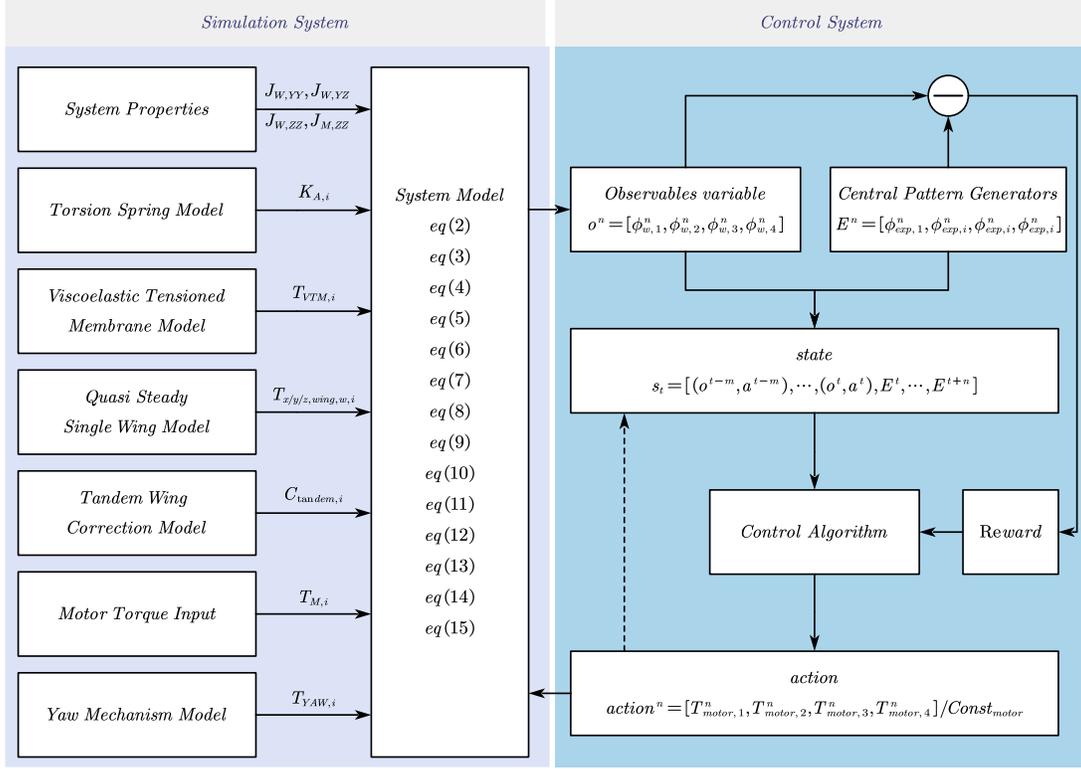

**Fig. 4. Simulation system and control system framework**

**Table 1 Description of the subsystem model**

| Subsystem | Function | Calculation method |
| --- | --- | --- |
| System Properties | Provide parameters such as wing inertia. | See literature for calculation method[21] |
| Torsion Spring Model | Provide spring stiffness. | See literature for calculation method[21] |
| Viscoelastic Tensioned Membrane Model | Provides motion constraints for four torsion angles. | See literature for calculation method[21] |
| Quasi Steady Single Wing Model | The single wing load part of the nonlinear aerodynamic load is modeled. | See literature for calculation method[21] |
| Tandem Wing Correction Model | Modeling the modification term of tandem wing model relative to single wing load. | See literature for calculation method[21] |
| Yaw Mechanism Model | Used to simulate the impact characteristics under yaw conditions. | See literature for calculation method[21] |

The motor torque input is the action output of the control algorithm. $T_{M,i}$ specifies the output torque of the i-th servo motor. $Const_{motor}$ is used to scale the action output in the range [−1, +1] to the actual required torque in N·m for the controlled object, as follows:

$$T_{M,1}, T_{M,2}, T_{M,2}, T_{M,3} = Const_{motor} \cdot action \tag{2}$$

Based on the above, the following 8-degree-of-freedom dynamic equation is obtained using the Lagrangian method[66]:

$$\ddot{\phi}_{w,1} = -\frac{J_{W,YY} \cdot K_{A,1}}{C_1} \cdot \phi_1 + \frac{J_{W,YY}}{C_1} \cdot T_{M,1} + \frac{J_{W,YY}}{C_1} \cdot T_{ZW,1} - \frac{J_{W,YZ}}{C_1} \cdot T_{YW,1} - \frac{J_{W,YZ}}{C_1} \cdot T_{VTM,1} - \frac{J_{W,YZ}}{C_1} \cdot T_{YAW,1} \tag{3}$$

$$\ddot{\phi}_{w,2} = -\frac{J_{W,YY} \cdot K_{A,2}}{C_1} \cdot \phi_2 - \frac{\pi \cdot J_{W,YY} \cdot K_{A,2}}{C_1} + \frac{J_{W,YY}}{C_1} \cdot T_{M,2} + \frac{J_{W,YY}}{C_1} \cdot T_{ZW,2} - \frac{J_{W,YZ}}{C_1} \cdot T_{YW,2}$$
$$- \frac{J_{W,YZ}}{C_1} \cdot T_{VTM,2} - \frac{J_{W,YZ}}{C_1} \cdot T_{YAW,2} \tag{4}$$

$$\ddot{\phi}_{w,3} = -\frac{J_{W,YY} \cdot K_{A,3}}{C_1} \cdot \phi_3 - \frac{\pi \cdot J_{W,YY} \cdot K_{A,3}}{C_1} + \frac{J_{W,YY}}{C_1} \cdot T_{M,3} + \frac{J_{W,YY}}{C_1} \cdot T_{ZW,3} - \frac{J_{W,YZ}}{C_1} \cdot T_{YW,3}$$
$$- \frac{J_{W,YZ}}{C_1} \cdot T_{VTM,3} - \frac{J_{W,YZ}}{C_1} \cdot T_{YAW,3} \tag{5}$$

$$\ddot{\phi}_{w,4} = -\frac{J_{W,YY} \cdot K_{A,4}}{C_1} \cdot \phi_4 + \frac{J_{W,YY}}{C_1} \cdot T_{M,4} + \frac{J_{W,YY}}{C_1} \cdot T_{ZW,4} - \frac{J_{W,YZ}}{C_1} \cdot T_{YW,4} - \frac{J_{W,YZ}}{C_1} \cdot T_{VTM,4}$$
$$- \frac{J_{W,YZ}}{C_1} \cdot T_{YAW,4} \tag{6}$$

$$\ddot{\theta}_{w,1} = \frac{J_{M,ZZ}}{C_1} \cdot T_{VTM,1} + \frac{J_{M,ZZ}}{C_1} \cdot T_{YW,1} + \frac{J_{W,YZ}}{C_1} \cdot K_{A,1} \cdot \phi_1 - \frac{J_{W,YZ}}{C_1} \cdot T_{M,1} - \frac{J_{W,YZ}}{C_1} \cdot T_{ZW,1} + \frac{J_{W,ZZ}}{C_1}$$
$$\cdot T_{VTM,1} + \frac{J_{W,ZZ}}{C_1} \cdot T_{YW,1} + \frac{J_{W,YZ}}{C_1} \cdot T_{YAW,1} \tag{7}$$

$$\ddot{\theta}_{w,2} = \frac{J_{M,ZZ}}{C_1} \cdot T_{VTM,2} + \frac{J_{M,ZZ}}{C_1} \cdot T_{YW,2} + \frac{J_{W,YZ} \cdot K_{A,2}}{C_1} \cdot \phi_2 + \frac{\pi \cdot J_{W,YZ} \cdot K_{A,2}}{C_1} - \frac{J_{W,YZ}}{C_1} \cdot T_{M,2}$$
$$- \frac{J_{W,YZ}}{C_1} \cdot T_{ZW,2} + \frac{J_{W,ZZ}}{C_1} \cdot T_{VTM,2} + \frac{J_{W,ZZ}}{C_1} \cdot T_{YW,2} + \frac{J_{W,YZ}}{C_1} \cdot T_{YAW,2} \tag{8}$$

$$\ddot{\theta}_{w,3} = \frac{J_{M,ZZ}}{C_1} \cdot T_{VTM,3} + \frac{J_{M,ZZ}}{C_1} \cdot T_{YW,3} + \frac{J_{W,YZ} \cdot K_{A,3}}{C_1} \cdot \phi_3 + \frac{\pi \cdot J_{W,YZ} \cdot K_{A,3}}{C_1} - \frac{J_{W,YZ}}{C_1} \cdot T_{M,3}$$
$$- \frac{J_{W,YZ}}{C_1} \cdot T_{ZW,3} + \frac{J_{W,ZZ}}{C_1} \cdot T_{VTM,3} + \frac{J_{W,ZZ}}{C_1} \cdot T_{YW,3} + \frac{J_{W,YZ}}{C_1} \cdot T_{YAW,3} \tag{9}$$

$$\ddot{\theta}_{w,4} = \frac{J_{MZZ}}{C} \cdot T_{VTM,4} + \frac{J_{M,ZZ}}{C} \cdot T_{YW,4} + \frac{J_{W,YZ}}{C} \cdot K_{A,4} \cdot \phi_4 - \frac{J_{W,YZ}}{C} \cdot T_{M,4} - \frac{J_{W,YZ}}{C} \cdot T_{ZW,4} + \frac{J_{W,ZZ}}{C}$$
$$\cdot T_{VTM,4} + \frac{J_{W,ZZ}}{C} \cdot T_{YW,4} + \frac{J_{W,YZ}}{C} \cdot T_{YAW,4} \tag{10}$$

$$C = J_{M,ZZ} \cdot J_{W,YY} + J_{W,YY} \cdot J_{W,ZZ} - J_{W,YZ}^2 \tag{11}$$

$$\begin{bmatrix} 0 \\ T_{YW,1} \\ T_{ZW,1} \end{bmatrix} = \begin{bmatrix} 0 \\ T_{y,\text{wing},w,1} \\ T_{z,\text{wing},w,1} \end{bmatrix} \cdot (C_{tandem,1} + 1) \tag{12}$$

$$\begin{bmatrix} 0 \\ T_{YW,2} \\ T_{ZW,2} \end{bmatrix} = \begin{bmatrix} 0 \\ T_{y,\text{wing},w,2} \\ T_{z,\text{wing},w,2} \end{bmatrix} \cdot (C_{tandem,2} + 1) \tag{13}$$

$$\begin{bmatrix} 0 \\ T_{YW,3} \\ T_{ZW,3} \end{bmatrix} = \begin{bmatrix} 0 \\ T_{y,\text{wing},w,3} \\ T_{z,\text{wing},w,3} \end{bmatrix} \cdot (C_{tandem,3} + 1) \tag{14}$$

$$\begin{bmatrix} 0 \\ T_{YW,4} \\ T_{ZW,4} \end{bmatrix} = \begin{bmatrix} 0 \\ T_{y,\text{wing},w,4} \\ T_{z,\text{wing},w,4} \end{bmatrix} \cdot (C_{tandem,4} + 1) \tag{15}$$

It is noteworthy that the spring stiffness $K_{A,i}$ is selected based on different flapping frequencies and maintained constant during operation to highlight the distinct characteristics of each environment. This adjustment principle underscores that the key structural parameters of the controlled object corresponding to different flapping frequency configurations also vary, as detailed in Section 5.2.

$$K_{A,i} = 4\pi^2 \cdot f_{exp}^2 \cdot J_{M,ZZ} \tag{16}$$

### C. Designing Reward Functions and Simulation Episodes

To facilitate the construction of a Lyapunov function[67] for proving algorithm convergence, a weighted sum of the absolute values of tracking errors for four motors, as shown in equation (17), is employed as the reward function. The goal is to achieve a minimum value under optimal performance:

$$R_t(s_t, a_t) = \lambda \cdot [|E_1(s_t, a_t)| + |E_2(s_t, a_t)| + |E_3(s_t, a_t)| + |E_4(s_t, a_t)|] \tag{17}$$

where $E_1(s_t, a_t)$, $E_2(s_t, a_t)$, $E_3(s_t, a_t)$, $E_4(s_t, a_t)$ denote the tracking errors of the four motors. The weighting factors $\lambda$ are used to scale the reward within the range [0, 1] to accelerate convergence[46].

In the context of simulator episode design, since this work focuses on direct-drive systems without mechanical limits, if the wing exceeds the maximum range of -90 to 90 degrees, it indicates that the wing has crossed the symmetry plane of the experimental platform and collided with the support structure. Therefore, tracking errors exceeding 90 degrees are presumed to inflict damage on the equipment.

## IV. Primarily

### A. On-Policy Deterministic Actor-Critic

The deterministic actor-critic[68] consists of two components. The critic estimates the action-value function while the actor ascends the gradient of the action-value function. Specifically, an actor adjusts the parameters $\theta$ of the deterministic policy $\mu$ by gradient ascent. A differentiable action-value function $Q(w|s, a)$ in place of the true action-value function $Q(\mu|s, a)$ is substituted. A critic estimates the action-value function $Q(w|s, a) \approx Q(\mu|s, a)$, using an appropriate policy evaluation algorithm. For example, in the following deterministic actor-critic algorithm, the critic uses Sarsa updates to estimate the action-value function[68, 69].

$$\delta = r_k + \gamma \cdot Q(w_k|s_{k+1}, a_{k+1}) - Q(w_k|s_k, a_k) \tag{18}$$

$$w_{k+1} = w_k + \alpha_w \cdot \delta \cdot \nabla_w Q(w_k|s_k, a_k) \tag{19}$$

$$\theta_{k+1} = \theta_k + \alpha_\theta \cdot \nabla_\theta Q(w_k|S_k, \mu(\theta_k|s_k)) \tag{20}$$

$$a_k = \mu(\theta_k|s_k) \tag{21}$$

where $Q(w|s, a)$ is the differentiable action-value function with network weight parameters $w$, and its value at the k-th time step is $w_k$. $\mu(\theta|S)$ is the policy network with weight parameters $\theta$, and its value at the k-th time step is $\theta_k$. $\gamma$ is the discount factor. $\alpha_w$ is the learning rate for parameter $w$. $\alpha_\theta$ is the learning rate for parameter $\theta$. $S_k$ is the visited state at the k-th time step.

### B. Q-Value Estimation Method Based on Gradient-Domain Laplace Transform

To estimate Q-values in real time without introducing a critic and to support the design and verification of the subsequent Rule-Based Policy Composer while considering finite-time impact and single-life impact, a Q-value estimation method based on the gradient-domain Laplace transform is adopted, as referenced in[21].

The gradient-domain Laplace transform operates in the $\theta$ domain, converting $\theta$ domain functions into functions of a complex variable $g$. The transformation principles are outlined in equations (22) to (25):

$$\theta = \int_0^\infty g \tag{22}$$

$$gd = SP + i \cdot DR \tag{23}$$

$$K(gd) = \int_0^\infty Q(\theta) \cdot e^{-s\theta} d\theta \tag{24}$$

$$R(gd) = \int_0^\infty r(\theta) \cdot e^{-s\theta} d\theta \tag{25}$$

Where $SP$ represents the real part of the gradient-domain Laplace transform, indicating the influence of the current gradient sequence update in the finite-time single-life task, analogous to the decay term in the Laplace transform. $DR$ represents the direction vector of the updated gradient, analogous to the frequency in the Laplace transform, indicating critical characteristics of the update process.

The gradient-domain Laplace transform shares similar properties with the classical Laplace transform. Using these properties, $K(grad)$ and $R(grad)$ are defined as:

$$K(gd) = \int_0^\infty Q(\theta) \cdot e^{-s\theta} d\theta \tag{26}$$

$$R(gd) = \int_0^\infty r(\theta) \cdot e^{-s\theta} d\theta \tag{27}$$

At the convergence point, for any given $\theta$, based on the Bellman equation[69], the following relationship holds:

$$Q_i(\theta) = r_i(\theta) + \gamma \cdot Q_{i+1}(\theta) \tag{28}$$

Taking the derivative with respect to $\theta$ on both sides:

$$\frac{dQ_i(\theta)}{d\theta} = \frac{dr_i(\theta)}{d\theta} + \gamma \cdot \frac{dQ_{i+1}(\theta)}{d\theta} \tag{29}$$

Applying the gradient-domain Laplace transform as defined above, the following is obtained:

$$g \cdot K_i(gd) - Q_i(0) = g \cdot R_i(g) - r_i(0) + \gamma \cdot K_{i+1}(gd) - Q_{i+1}(0) \tag{30}$$

Simplifying this yields:

$$K_i(g) = \frac{R_i(gd)}{1 - \gamma} \tag{31}$$

Performing the inverse gradient Laplace transform yields the following relationship, thus establishing an approximate connection between $r_i(\theta)$ and $Q_i(\theta)$, Based on the gradient-domain Laplace transform, $Q_i$ values are estimated to some extent. However, a significant issue in the above estimation is that $\gamma_i$ essentially represents the confidence distribution of $r_i(\theta)$ with respect to $Q_i(\theta)$. To mitigate the error in estimating $Q_i(\theta)$ based on $r_i(\theta)$:

$$Q_i(\theta) = \frac{r_i(\theta)}{1 - \gamma_i} \tag{32}$$

Assumption 1: By selecting an appropriate $L$, $\gamma_i$ can be approximated by $\gamma_{mean}$, the mean value of $\gamma_i$ over the entire single-life process, i.e., $\gamma_i \approx \bar{\gamma}$.

Evidence: Existing research indicates that when the single-life duration is sufficiently long, the influence of the current gradient sequence update in the finite-time single-life task becomes negligible[70, 71].

In practice, based on Assumption 1, equation (32) can be averaged over L time steps for $Q$ and $r$, yielding:

$$\bar{Q}(\theta) \approx \frac{\bar{r}(\theta)}{1 - \bar{\gamma}} \tag{33}$$

From equation (33), the following inference allows the calculation of $\Delta Q$ under the conditions $\theta + \Delta\theta$.

$$\Delta\bar{Q}(\Delta\theta) \approx \frac{\bar{r}(\theta + \Delta\theta) - \bar{r}(\theta)}{1 - \bar{\gamma}} \tag{34}$$

## V. The proposed algorithm: CRL2RT algorithm

### A. Architecture of the CRL2RT algorithm

The architecture of the CRL2RT algorithm, as illustrated in Fig.5 primarily comprises a Cloud side for updating neural network weights and an Edge side for inference. On the Edge side, a load-balancing approach[72] is employed, defining Mode 1: the integration of traditional control components with the Reception, Analysis, and Loading of Weight modules and the CRL algorithm's traditional control components and Mode 2: learning based control. These are executed in parallel with the conventional reinforcement learning inference components, operating in an interleaved fashion based on the Time-Interleaved mode.

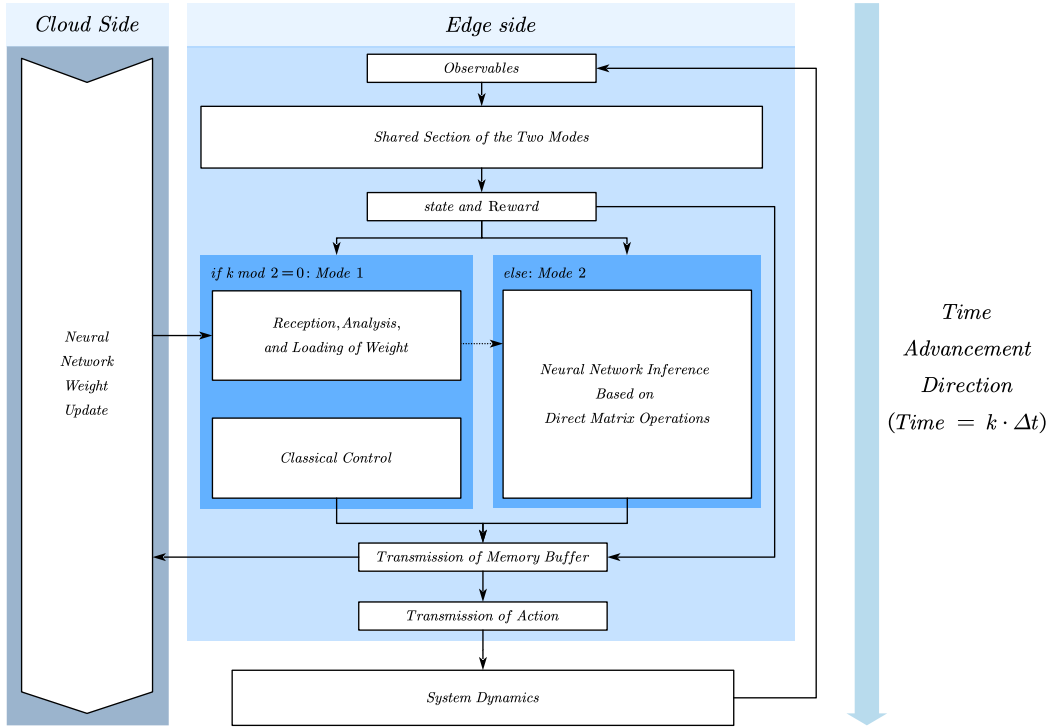

**Fig. 5. CRL2RT algorithm's architecture**

Based on this framework, the mathematical expressions of the CRL2RT are divided into two parts. On the Edge side, the following operations are performed:

$$if\ k\ mod\ 2 = 0: \tag{35}$$

$$a_k = \mu_{classical}(s_k) \tag{36}$$

$$\theta_k \rightarrow \hat{\theta}_{ethnet,k-1} \rightarrow \theta_{edge,k-1} \tag{37}$$

$$else: \tag{38}$$

$$action_k = \mu_{RL}(\theta_{edge}|s_k) \tag{39}$$

On the Cloud side, the following operations are executed:

$$a_k = \mu_{RL}(\theta_k|s_k) \tag{40}$$

$$\delta = r_k + \gamma \cdot Q(w_k|s_{k+1}, a_{k+1}) - Q(w_k|s_k, a_k) \tag{41}$$

$$w_{k+1} = w_k + \alpha_w \cdot \delta \cdot \nabla_w Q(w_k|s_k, a_k) \tag{42}$$

$$\theta_{k+1} = \theta_{CP,j,g}(\theta, t) + \theta_k + \Delta\theta_{critic,k} \tag{43}$$

$$\Delta\theta_{critic,k} = -\alpha_\theta \cdot \nabla_\theta Q(w_k|S_k, \mu(\theta_k|s_k)) \tag{44}$$

$$\frac{dS}{dt} = \frac{F_{explore}(t)}{\Delta t} \tag{45}$$

Compared to the previously mentioned On-Policy Deterministic Actor-Critic, the following:
1. $\mu_{classical}(s_k)$ indicates the classical controller.
2. $\mu_{RL}(\theta_k|s_k)$ indicates the reinforcement learning controller.
3. $if\ k\ mod\ 2 = 0$, the Time-Interleaved module alternates the execution of the classical controller $\mu_{classical}(s_k)$ and the reinforcement learning controller $\mu_{RL}(\theta_k|s_k)$.
4. $\theta_{CP,j,g}(\theta, t)$ represents the additional policy weight introduced by the Rule-based Policy Composer in the $j$-th gradient descent segment of the $g$-th dynamic descent phase. Detailed functionality is described in Section 4.3.
5. $\theta_{init,g}$ denotes the initial network weight for the $g$-th dynamic descent phase. This is the initial network weight for the first segment. Subsequently, $\theta_{init,g}$ adopts the weight $\theta_{CP,j}(\theta, t)$, i.e., $\theta_{init,j} = \theta_{init,j-1} + \theta_{CP,j}(\theta, t)$. Details are elaborated in Section 4.3.
6. The reward is defined to achieve its minimum value at optimal performance to facilitate the Lyapunov function-based stability proof. The calculation of $\theta_{k+1}$ uses $-\alpha_\theta \cdot \nabla_\theta Q(w_k|S, \mu(\theta_k|s_k))$, with a negative sign.
7. $F_{explore}(t)$ is introduced as the exploration strategy, derived from the Rule-based Policy Composer's adjustment process and the classical controller's control process from the Time-Interleaved module. This enables the integration of data from both processes, along with the system dynamics interaction, into the replay buffer.
8. $\theta_k \rightarrow \hat{\theta}_{ethnet,k-1} \rightarrow \theta_{edge,k-1}$ represents the Reception, Analysis, and Loading of Weight module, indicating the network update process under algorithmic delay and network update lag conditions.

B. **Design of Time-Interleaved module and Guarantee of Lipschitz Conditions**

The primary function of the Time-Interleaved module is to enhance overall process safety by alternating the execution of classical controllers, which have basic feedback control capabilities, with reinforcement learning controllers. The pseudo-code is presented below.

---
**Algorithm 1:** Time-interleaved Mechanism
---
1  for $k \leftarrow 0$ to $N$ do
2     if $k\ \mathrm{mod}\ 2 = 0$ then
3        $a_k \leftarrow \mu_{\mathrm{classical}}(s_k)$
4     else
5        $a_k \leftarrow \mu_{\mathrm{RL}}(\theta_k\ |\ s_k)$
6     end
7  end
---

The theoretical basis for the functionality of the Time-Interleaved module and Time-Interleaved Capability is explained using Lipschitz Conditions:

First, a vector $\overrightarrow{Er}(t)$ representing the tracking error of the four wings relative to the desired commands over time is defined:

$$\overrightarrow{Er}(t) = [E_1(t), E_2(t), E_3(t), E_4(t)] \tag{46}$$

For two adjacent time steps, as defined by the Time-Interleaved module (refer to Algorithm 1), which alternates between the RL controller and the classical controller, the error generated between these two time steps are:

$$\overrightarrow{Er}(t + \Delta t) = \overrightarrow{Er}(t) + \overrightarrow{PE}_{rl}(t) \cdot \Delta t \tag{47}$$

$$\overrightarrow{Er}(t + 2 \cdot \Delta t) = \overrightarrow{Er}(t + \Delta t) - \overrightarrow{PD}_{class} \cdot \Delta t \tag{48}$$

Here, $\overrightarrow{PE}_{rl}(t)$ represents the RL algorithm's ability to increase or decrease the error, and $\overrightarrow{PC}_{class}$ represents Time-Interleaved Capability, i.e., the classical control algorithm's ability to reduce the error, which remains constant.

The relationship can be expressed as:

$$\overrightarrow{Er}(t + 2 \cdot \Delta t) - \overrightarrow{Er}(t) = -\overrightarrow{PC}_{class} \cdot \Delta t + \overrightarrow{PE}_{rl}(t) \cdot \Delta t \tag{49}$$

Assuming $\Delta t^* = 2 \cdot \Delta t$ and the maximum value of $\overrightarrow{PE}_{rl}(t)$ is $\overrightarrow{PE}_{rl,max}$, the following inequality is established:

$$\overrightarrow{Er}(t + \Delta t^*) - \overrightarrow{Er}(t) = \frac{-\overrightarrow{PC}_{class} + \overrightarrow{PE}_{rl}(t)}{2} \Delta t^* < \frac{-\overrightarrow{PC}_{class} + \overrightarrow{PE}_{rl,max}}{2} \Delta t^* \tag{50}$$

This can be expressed in terms of Lipschitz continuity as follows:

$$\Delta \overrightarrow{Er} = \overrightarrow{Er}(t + \Delta t^*) - \overrightarrow{Er}(t) < \lambda_{\text{Lipschitz}} \cdot \Delta t^* \tag{51}$$

$$\lambda_{\text{Lipschitz}} = \frac{-\overrightarrow{PC}_{class} + \overrightarrow{PE}_{rl,max}}{2} \tag{52}$$

Therefore, the Time-Interleaved module, with its Time-Interleaved Capability, ensures that the algorithm satisfies Lipschitz continuity. Based on the above equation, the following analysis can be made:

1. Adjustable Range of $\Delta \overrightarrow{Er}$: According to equation (51), the desired $\Delta \overrightarrow{Er}$ can be achieved by adjusting the relative relationship between $\overrightarrow{PC}_{class}$ (i.e., Time-Interleaved Capability) and $\overrightarrow{PE}_{rl,max}$ to meet hard safety constraints. For example, if $\Delta \overrightarrow{Er} < \Delta \overrightarrow{Er}_{max}$, then $\overrightarrow{PC}_{class}$ must satisfy the following conditions:

$$\overrightarrow{PC}_{class} > \overrightarrow{PE}_{rl,max} - \frac{\Delta \overrightarrow{Er}_{max}}{\Delta t} \tag{53}$$

2. Critical Point for Switching $\overrightarrow{PC}_{class}$: When $\overrightarrow{PC}_{class} = \overrightarrow{PE}_{rl}(t)$, this represents a critical point ensuring boundedness and guaranteeing a gradual reduction of the error ($\Delta \overrightarrow{Er} < 0$). If $\overrightarrow{PC}_{class} < \overrightarrow{PE}_{rl}(t)$, equation (53) transforms into the form of equation (54), where the influence of $\overrightarrow{PC}_{class}$ on $\overrightarrow{PE}_{rl}(t)$ must be accounted for to mitigate adverse effects and enhance performance:

$$\frac{\Delta \overrightarrow{Er}}{\Delta t} + \overrightarrow{PC}_{class} = \overrightarrow{PE}_{rl}(t) \tag{54}$$

To incorporate perturbations into the Time-Interleaved Capability perturbation (TICP) is applied as described below:

$$\overrightarrow{PC}_{class} = \overrightarrow{PC}_{class,O} \pm \varepsilon_{TICP} \tag{55}$$

Here, $\overrightarrow{PC}_{class,O}$ represents the original Time-Interleaved Capability, while $\varepsilon_{TICP}$ denotes the perturbation induced by changes in the classical controller's capability. Notably, the Time-Interleaved Capability Perturbation is directly implemented by modifying the classical controller's strategy during the inference phase, making it suitable for real-time execution.

### C. Design of Rule-Based Policy Composer and Guarantee of Finite-Time Lyapunov Convergence Conditions
#### 1) Time Segmentation of Single-Life and Training Phase Definitions

This study employs an approach designed to suppress erroneous gradient updates to enhance training efficiency. Given the unidirectional and unique nature of single-life processes, a rule is established to record the policy parameters, $\theta$, during the single-life progression.

Building on the gradient-domain Laplace transform concept introduced in Section 3.2 and addressing the requirements of the Rule-Based Policy Composer, the single-life timeline is segmented into two hierarchical training phases, as depicted in Fig.6. Specifically, every L time step constitutes a gradient descent segment (GDS). At the same time, up to every N of the GDSs form a dynamic descent phase (DDP).

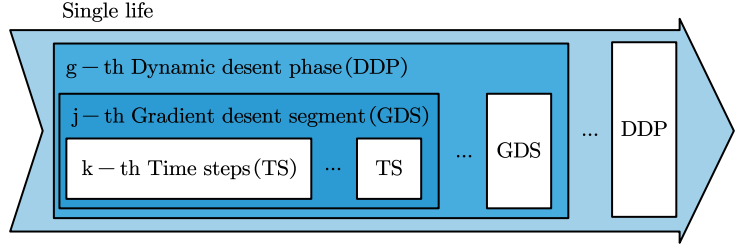

**Fig. 6 Schematic diagram of two hierarchical training phases**

Based on the segmentation of these two training phases, the following phase-specific policy network weight parameters are defined: $\theta_{g,j}(\theta, t)$, representing the initial policy weight for the $j$-th GDS in the g-th DDP. Additionally, the gradient update parameters $\Delta\theta_j$ for each GDS and $\Delta\theta_g$ for each DDP are defined as follows:

$$\Delta\theta_j = \sum_{k=0}^{L} \Delta\theta_{critic,k} \tag{56}$$

$$\Delta\theta_g = \sum_{j=0}^{N} \Delta\theta_j \tag{57}$$

**2) Proof of Finite-Time Lyapunov Convergence Conditions**

The primary function of the Policy Composer at the beginning of each GDS is to introduce an additional posterior-determined weight, $\theta_{CP,g,j}(\theta, t)$, based on existing policy weights. This adjustment enhances the algorithm's convergence speed and effectiveness during control tasks. The calculation method for $\theta_{CP,g,j}(\theta, t)$ and its impact on the algorithm's convergence are as follows:

Referring to equation (34), $\theta_{CP,g,j}(\theta, t)$ is defined as the neural network weight satisfying the following posterior acceptance condition. This condition ensures that $\theta_{CP,g,j}(\theta, t)$ accelerates the decrease in $\overline{\overline{Q_g}}(\Delta\theta)$ over progressively shorter time spans within each dynamic descent phase (DDP), which is expressed in the equation (59):

$$\overline{\overline{Q_g}}(\Delta\theta) = \frac{\sum_0^N \overline{Q_{g,j}}(\Delta\theta)}{N} \tag{58}$$

$$\frac{\overline{\overline{Q_g}}(\Delta\theta)}{\Delta\theta} = (1 + m_g) \cdot \frac{\overline{\overline{Q_{g-1}}}(\Delta\theta)}{\Delta\theta} \tag{59}$$

Assuming that an initial $\overline{\overline{Q_0}}(\Delta\theta)$ exists at the beginning of the single-life process, the following expression is derived:

$$\frac{\overline{\overline{Q_g}}(\Delta\theta)}{\Delta\theta} = \prod_0^g (1 + m_g) \cdot \frac{\overline{\overline{Q_0}}(\Delta\theta)}{\Delta\theta} \tag{60}$$

The primary function of the Policy Composer is to achieve the following posterior-screened expression:

$$\frac{d\theta_{CP,g,j}(t)}{dt} = \frac{\overline{\overline{Q_g}}(\Delta\theta)}{\Delta\theta} = \prod_0^g (1 + m_g) \cdot \frac{\overline{\overline{Q_0}}(\Delta\theta)}{\Delta\theta} \tag{61}$$

Combining with the earlier equations, the relationship becomes:

$$\frac{d\theta}{dt} = \prod_0^g (1 + m_g) \cdot \frac{\overline{\overline{Q_0}}(\Delta\theta)}{\Delta\theta} - \alpha_\theta \cdot \frac{dQ}{d\theta} \tag{62}$$

Based on the Rule-Based Policy Composer form $\theta_{CP,j,g}(\theta, t)$ obtained above, the subsequent sections will demonstrate its impact on the convergence of the entire algorithm. To ensure convergence within a finite time, the selected Lyapunov function, $Q$, must satisfy the following finite-time Lyapunov convergence condition[73]:

$$\frac{dQ(t)}{dt} \leq -c \cdot Q(t)^\alpha \tag{63}$$

For the convenience of using continuous functions in the convergence proof, $Q$ is considered as a function of $w$, $S$, $\mu$ and $\theta$. Differentiating $Q$ yields:

$$\dot{Q}(w(t), S(t), \mu(\theta(t), S(t))) = \frac{dQ}{dt} = \frac{dQ}{dw} \cdot \frac{dw}{dt} + \frac{dQ}{dS} \cdot \frac{dS}{dt} + \frac{dQ}{d\mu} \cdot \frac{d\mu}{dt} \tag{64}$$

Based on the Critic update method defined in equation (41), it can be transformed into the following form:

$$\frac{d\mu}{dt} = \frac{d\mu}{d\theta} \cdot \frac{d\theta}{dt} \quad (65)$$

$$\frac{dw}{dt} = +\frac{\alpha_w \cdot \delta \cdot \frac{dQ}{dw}}{\Delta t} \quad (66)$$

$$\frac{dS}{dt} = \frac{F_{explore}(t)}{\Delta t} > 0 \quad (67)$$

$$\frac{dQ}{dS} < 0 \quad (68)$$

$$\frac{dQ}{d\theta} < 0 \quad (69)$$

$$\delta = r_k + \gamma \cdot \left(\frac{dQ}{dS} \cdot \frac{dS}{dt} + \frac{dQ}{d\theta} \cdot \frac{d\theta}{dt}\right) \cdot \Delta t \quad (70)$$

$$[r_k + (\gamma - 1) \cdot Q] \to 0 \quad (71)$$

Substituting these into the time derivative of $Q(t)$, the result is:

$$\begin{aligned}\frac{dQ(t)}{dt} &= \left[\left(\frac{dQ}{dw}\right)^2 \cdot \frac{\alpha_w}{\Delta t} \cdot \gamma + 1\right] \cdot \frac{dQ}{dS} \cdot \frac{F_{explore}(t)}{\Delta t} \cdot \Delta t + \left[\left(\frac{dQ}{dw}\right)^2 \cdot \alpha_w \cdot \gamma + 1\right] \cdot \frac{dQ}{d\theta} \\ &\cdot \prod_0^g (1 + m_g) \cdot \frac{\overline{\overline{Q_0}}(\Delta\theta)}{\Delta\theta} - \left[\left(\frac{dQ}{dw}\right)^2 \cdot \alpha_w \cdot \gamma + 1\right] \cdot \frac{dQ}{d\theta} \cdot \alpha_\theta \cdot \frac{dQ}{d\theta}\end{aligned} \quad (72)$$

To satisfy the finite-time Lyapunov convergence condition:

$$\frac{dQ}{d\theta} \cdot \prod_0^g (1 + m_g) \cdot \frac{\overline{\overline{Q_0}}(\Delta\theta)}{\Delta\theta} \leq -c^* \cdot Q(t)^\alpha \quad (73)$$

$$c^* = \frac{c}{\left[\left(\frac{dQ}{dw}\right)^2 \cdot \alpha_w \cdot \gamma + 1\right]} \quad (74)$$

From equation (73), it is evident that $\prod_0^g(1 + m_g)$ grows exponentially, while $Q(t)^\alpha$ grows polynomially. Thus, the finite-time Lyapunov convergence condition is satisfied.

**3) Physics-Inspired Rule-Based Policy Composer strategy**

A feasible approach to implement the strategy described by equation (59) involves decomposition at the g-th DDP level based on whether performance has been effectively updated or remains unchanged due to timeout. Subsequently, at the j-th GDS level, the Q-variation within the j-th GDS is linearly fitted to time steps according to the Gradient-Domain Laplace Transform theory. This fitting yields the descent slope and descent intercept. However, relying solely on the mean descent intercept proves effective only under constant task descent speeds and is inadequate under pitch, roll, and yaw operating conditions.

To address this limitation, it is essential to evaluate the generalization performance and adaptability of the algorithm under pitch, roll, and yaw operating conditions. Therefore, a standard deviation $c_j$ of the relative average descent performance for each time step is introduced as a metric. The expression is given by:

$$c_j = \sqrt{\frac{\sum_{i=0}^L \left(Q_{g,j,k} - (a_j \cdot k + b_j)\right)^2}{L}} \quad (75)$$

Building upon this, policy optimization is interpreted from a Physics-Inspired perspective as the state evolution in a dynamic system. Specifically, the descent slope $a_j$ is analogous to potential energy, while the standard deviation $c_j$ represents kinetic energy.

Additionally, drawing inspiration from the composer perturbation introduced in biological memory reorganization processes[53, 74], a noise term $\theta_{noise}(x, \theta_k)$ representing random noise with a mean of zero and a standard deviation of scaled by $\theta_k$ is incorporated. Based on this formulation, the following pseudocode is proposed:

**Algorithm 2:** Physics-Inspired Rule-Based Policy Composer Strategy

```
1  if \overline{\overline{Q}}_{g,j}(\Delta\theta) > \overline{\overline{Q}}_{g,0}(\Delta\theta) then
       // Exceed the current historical optimal: Receive all
          updates
2      θ_{g+1,0}(θ,t) ← θ_j + Σ_{k=0}^{L} Δθ_{critic,k}
3      j ← 0
4  else
5      if j > N then
           // Not updated for a long time: Reset the gradient
6          θ_{g+1,0}(θ,t) ← θ_{g,0}(θ,t)
7          j ← 0
8      else
9          j = j + 1
10         if a_{j-1} < a_j then
11             if c_{j-1} < c_j then
                   // Expected state: introduce all update
                      gradients
12                 θ_{j+1} ← θ_j + Σ_{k=0}^{L} Δθ_{critic,k}
13             else
                   // Expected state but large variance: Reset
                      partial gradient
14                 θ_{j+1} ← θ_j + β · Σ_{k=0}^{L} Δθ_{critic,k} + θ_{noise}(0.064, θ_k)
15             end
16         else
17             if c_{j-1} < c_j then
                   // Slowing down: removes the update gradient and
                      adds random noise for exploration
18                 θ_{j+1} ← θ_j + θ_{noise}(0.04, θ_k)
19             else
                   // Update failure: Explore around the current
                      best
20                 θ_k ← θ_{g,0} + θ_{noise}(0.2, θ_k)
21             end
22         end
23     end
24 end
```

Here, $\beta \in [0,1]$ represents a scaling factor. It is worth noting that the above process primarily applies to the training phase and involves substantial computations, thus restricting its implementation to training devices. As a result, composer perturbation is also executed on training devices and is not suitable for real-time application.

## D. Defining the State Spaces and Constructing the Networks and Real-Time Testing

The core challenge of the current problem lies in its inherently nonlinear and non-stationary nature, coupled with issues of partial observability[75]. Additionally, high-frequency control introduces typical high-frequency time series problems, including high volatility and nonlinearity, heterogeneity, strong noise and outliers, and non-stationarity[76, 77]. These factors impose higher requirements on the rationality of state design.

Based on Takens' theorem, the system's state space can be reconstructed by selecting an appropriate embedding dimension, observing a single variable, and performing delay embedding on the generated time series data. The embedded trajectory maintains topological equivalence with the original system trajectory, and existing research has successfully validated the engineering guidance value of this theorem[65]. Following these studies, the state $s_t$ is designed to include as much information as possible from multiple past time steps and future commands, based on equation (76), with $M = 7$ and $H = 3$.

$$s_t = [(o^{t-M}, a^{t-M}), \cdots, (o^t, a^t), E^t, \cdots, E^{t+H}] \tag{76}$$

$$o^n = [\phi_{w,1}^n, \phi_{w,2}^n, \phi_{w,3}^n, \phi_{w,4}^n] \tag{77}$$

$$a^n = [T_{motor,1}^n, T_{motor,2}^n, T_{motor,3}^n, T_{motor,4}^n] \tag{78}$$

$$E^n = [\phi_{exp,1}^n, \phi_{exp,i}^n, \phi_{exp,i}^n, \phi_{exp,i}^n] \tag{79}$$

## E. Constructing the Networks

Given the high-frequency characteristics of the target system, addressing the plug-and-play, fully on-the-job, single-life, and lifelong non-stationarity problem necessitates rapid and continuous updates to network weights. An efficient network validated in existing research[21] is employed, as illustrated in Fig.7.

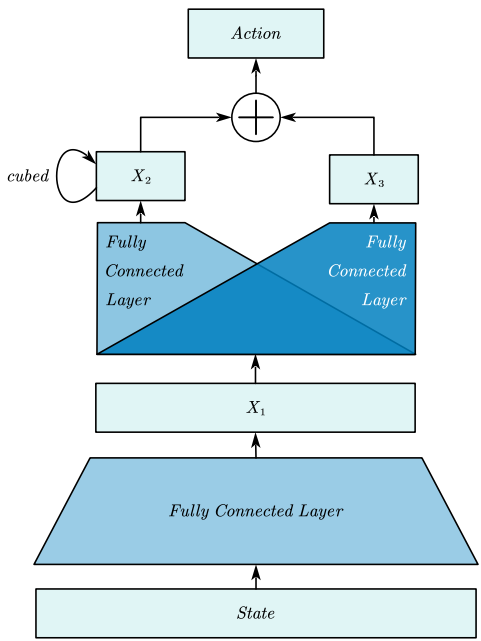

**Fig. 7. Actor network**

The critic network adopts a seven-layer MLP structure, as shown in Fig 8. In comparative algorithms like the SAC algorithm, Double Q-Learning is required, leading to the introduction of two identically structured but differently initialized critics. It is noteworthy that the critic is not used in edge computing.

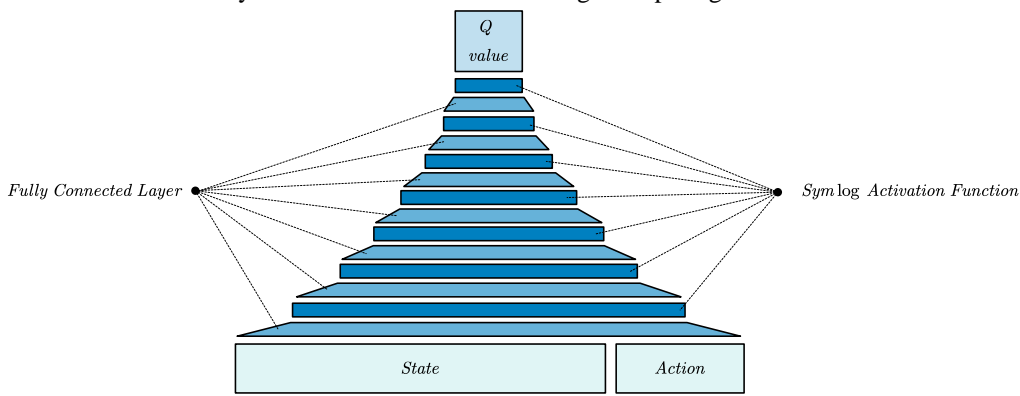

**Fig. 8. Critic network**

### F. Custom Edge Inference Framework for Real-Time Control

Existing studies typically employ PyTorch for the policy inference process in reinforcement learning[75, 78]. However, the real-time control at 2000 Hz emphasizes I/O overhead and the additional time required for converting received numpy array data to tensors. To address these challenges, the neural network inference process has been restructured into matrix operations, and an edge computing framework has been implemented using the NumPy library[79].

## VI. Experiments

### A. Objectives and Research Questions

This study aims to address the following questions through a series of experiments:

1. Question 1: What enhancements in real-time performance are achieved by the proposed algorithm and deployment scheme?
2. Question 2: How is the applicability of the CRL framework to various classical controllers evaluated under real hardware interference conditions?

### B. Experimental settings
#### 1) Basic Training and Simulation Parameters and Computing Devices

A uniform experimental setup was adopted for the algorithm comparisons conducted in this study. During the algorithm execution process, a semi-physical simulation system, as illustrated in Fig 9, was first constructed. The hardware and software configurations used in the experiments included Python version 3.12, PyTorch version 2.0, and Ubuntu 24.0 as the operating system. The cloud side component utilized a computer equipped with an Intel 13700KF CPU and an NVIDIA RTX 4090D GPU. The edge side employed an AMD 7840HS industrial controller, and the Physical Simulation Computer had an Intel 13700KF CPU. For simulation, the system model shown in Fig.4 was utilized. Due to the numerical stiffness caused by the rapid tensioning process in the system model[80], the Radau algorithm[81], an implicit numerical method, was adopted to ensure numerical stability. The interval for each event step during the iteration process was set to 0.0005 seconds.

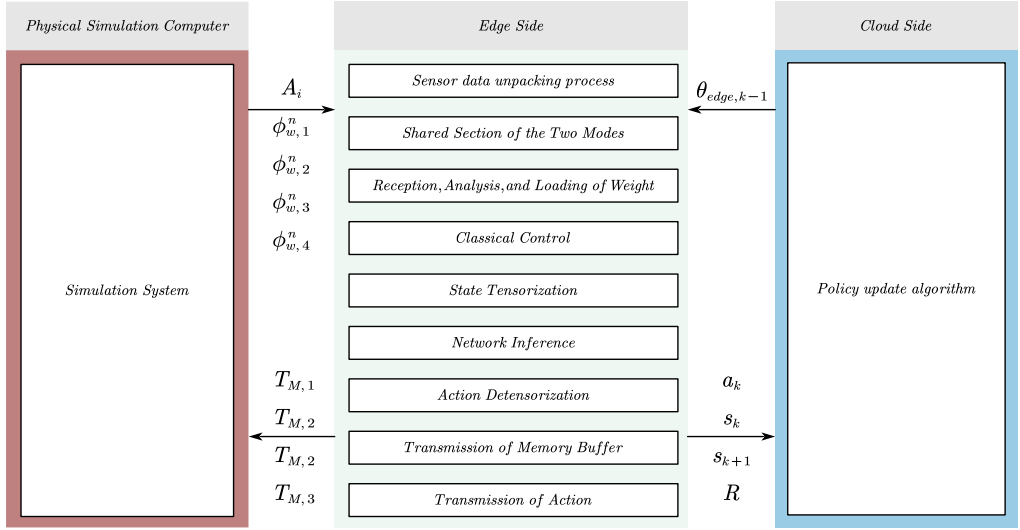

**Fig. 9. Schematic of the semi-physical simulation system**

In terms of training, the network structure and initial parameters of both the actor network and the critic network were kept consistent across all experiments. For network initialization, refer to Section 4.5 for the definition of the network. Each linear layer's weights were initialized using a Xavier uniform distribution[78] with a gain of 1, and the biases were initialized to zero.

All network parameters were optimized using the Lion optimizer, known for better convergence speeds compared to AdamW and other optimization algorithms across various tasks[82]. The learning rates were set at 0.015 for the actor and 0.0015 for the critic. The discount factor $\gamma$ for the critic was set to 0.9. Training for all algorithms commenced from the 10th step. The batch size for critic updates was set to 512, while that for policy updates was 16. The replay buffer size was set to 2000.

Regarding the scale of the experiments, as the theoretical convergence of the algorithm has already been discussed, the focus of the following experiments is on comparing convergence speeds within a limited number of time steps. The iteration count was chosen to be 450,000 steps[65].

#### 2) Classical Controller Parameters for Time-Interleaved module

Various classical controllers were employed in this study to investigate the generalizability of the CRL framework. The parameter settings for these controllers are presented below. It is important to note that these parameters primarily satisfy the requirements of the $\overrightarrow{PC}_{class}$ rather than being optimized control parameters.

The PID controller is a classic control mechanism that determines the control torque based on the current tracking error and relevant parameters. Following existing research, the PID controller parameters were selected as shown in Table 2:

**Table 2 Description of Control Parameters in the PID controller**

| Parameter | Value |
|---|---|
| $KP_{init}$ | $4.80 \times 10^{-1}$ |
| $KI_{init}$ | $2.00 \times 10^{-5}$ |
| $KD_{init}$ | $7.00 \times 10^{-4}$ |

The Adaptive PID controller updates its parameters based on the MIT principle[65]. To prevent the parameters from decreasing in later iterations and failing to meet the $\overrightarrow{PC}_{class}$ requirements, the parameters for the adaptive controller were 20% higher than those of the PID controller, with a learning rate of 0.000005, as shown in Table 3:

**Table 3 Description of Control Parameters in the Adaptive PID controller**

| Parameter | Value |
|---|---|
| $KP_{init}$ | $5.76 \times 10^{-1}$ |
| $KI_{init}$ | $2.40 \times 10^{-5}$ |
| $KD_{init}$ | $8.40 \times 10^{-4}$ |

Model Reference Adaptive Control (MRAC) aims to make the actual system output follow the reference model output, thereby achieving the desired dynamic response[83]. It is an adaptive control method that adjusts controller parameters in real time based on the system's operational state, commonly used for systems with uncertainties or time-varying parameters. The MRAC controller employed in this study utilizes an autoregressive model to address the online real-time finite-time single-life learning problem, and the feedback control adopts a PID controller, as detailed in Table 4:

Table 4 Description of Control Parameters in the Feedback Component of MRAC

| Parameter | Value |
|---|---|
| $KP_{init}$ | $5.76 \times 10^{-1}$ |
| $KI_{init}$ | $2.40 \times 10^{-5}$ |
| $KD_{init}$ | $8.40 \times 10^{-4}$ |

### 3) Operating Conditions Plan

Different frequencies and load types were considered to investigate the motion control issues of the DDTWEP under pitch, roll, and yaw load conditions for quad-wing direct-drive biomimetic aircraft.

This decision was based on the observation that direct-drive aircraft, such as the DDTWEP, often utilize springs and resonance to mitigate the torque required by the motors [17, 37, 84-87]. The selected frequencies of 20 Hz, 40 Hz, and 60 Hz were chosen based on the flapping frequency range of biological dragonflies and biomimetic aircraft [11], with the corresponding spring stiffness determined using Equation (16).

Regarding amplitude, the flapping amplitude of biological dragonflies is primarily 60°[11]. Therefore, a flapping amplitude of 60° ± 15° was selected. Specifically, the flapping amplitude was generated using a stochastic approach. With a 50% probability, the amplitude was set to a constant value of 60°. Otherwise, the amplitude was sampled from a predefined set of values [45, 47.5, 50, 52.5, 55, 57.5, 60, 62.5, 65, 67.5, 70, 72.5, 75] based on a normal distribution with a mean of 60 and a standard deviation of 10, ensuring higher probabilities for values closer to the mean. The desired torque was generated based on Equation (1). Additionally, to further explore the capabilities of the control algorithm, a yaw control mechanism was introduced as an additional load to create more challenging experimental conditions. This was implemented by adding an extra $T_{VTM,i}$ to Equations (2) to (15), with TVTM, $T_{VTM,i}$ set to 0 in conditions without the yaw mechanism.

In summary, the six operating conditions considered in this study are as follows in table 5.

**Table 5 Selected operating condition**

| Condition | $f/Hz$ | $A_{min}$ | $A_{max}$ | Whether introduce yaw mechanism |
|---|---|---|---|---|
| Load 1, 20 Hz configuration | 20 | 45° | 75° | No |
| Load 1, 40 Hz configuration | 40 | 45° | 75° | No |
| Load 1, 60 Hz configuration | 60 | 45° | 75° | No |
| Load 2, 20 Hz configuration | 20 | 45° | 75° | Yes |
| Load 2, 40 Hz configuration | 40 | 45° | 75° | Yes |
| Load 2, 60 Hz configuration | 60 | 45° | 75° | Yes |

In summary, six operating conditions were considered in this study, encompassing three different frequencies with and without the yaw control mechanism, ensuring a comprehensive evaluation under varying dynamics:

#### 4) Algorithms Participating in the Comparison

To address the research questions presented in Section 5.1, six algorithms were evaluated. These algorithms are listed in Table 6 below:

**Table 6 Overview of Algorithms Participating in the Comparison**

| Algorithm | Overview |
| --- | --- |
| 2000Hz-PID | Executes the PID controller at a frequency of 2000Hz. |
| 2000Hz-APID | Executes the APID controller at a frequency of 2000Hz. |
| 2000Hz-MRAC | Executes the MRAC controller at a frequency of 2000Hz. |
| CRL2RT(PID) | Implements the CRL2RT algorithm using PID as the classical controller in the Time-Interleaved module. |
| CRL2RT(APID) | Implements the CRL2RT algorithm using APID as the classical controller in the Time-Interleaved module. |
| CRL2RT(MRAC) | Implements the CRL2RT algorithm using MRAC as the classical controller in the Time-Interleaved module. |

### C. Question 1: What enhancements in real-time performance are achieved by the proposed algorithm and deployment scheme?

To deploy the controller in real-time control scenarios, a semi-physical simulation system, as illustrated in Fig 9 was constructed. The time required for each step in the single-step control process during the entire algorithm execution was measured and presented in Table 7 to Table 10.

**Table 7 Inference Process Time Series**

| Item | CRL2RT-Adjusted Framework-Direct Matrix Operation Inference | CRL2RT-Direct Matrix Operation Inference | CRL2RT |
| --- | --- | --- | --- |
| Sensor data unpacking process | | 9.06 E -06 | |
| Shared Section of the Two Modes/s | | 1.78E-04 | |
| Reception, Analysis, and Loading of Weight /s | 3.12E-04 | 3.12E-04 | 1.58E-07 |
| Classical Control/s | 1.69E-05 | 0.0 or 1.69 E-05 | |
| State Tensorization/s | 1.89E-04 | 0.0 or 1.89E-04 | 0.0 |
| Network inference/s | 3.01E-04 | 0.0 or 3.01E-04 | 0.0 or 4.78E-05 |
| Action de-tensorization/s | 2.39E-05 | 0.0 or 2.39E-05 | 0.0 |
| Total Algorithm Execution Time/s | 8.43E-04 | 3.29E-04~5.14E-04 | 1. 69E-05~4.78E-05 |
| Transmission of Memory Buffer/s | | 1.12E-04 | |
| Transmission of Action/s | | 4.78E-05 | |
| Slowest Execution Time/s | 1.50E-03 | 1.17E-03 | 3.95E-04 |
| Slowest Control Frequency /Hz | 665.87 | 852.66 | 2534.21 |

**Table 8 Shared Section of the Two Modes – Task Duration and Function**

| Task | Typical Duration (s) | Function |
| --- | --- | --- |
| CPG-based desired command generation | 1.04E-04 | Executes the central pattern generator oscillator as per Equation (1) |
| State generation | 4.54E-05 | Generates state based on angle and torque information as per Equation (76) |
| New-old alternation 1 | 1.56E-06 | Intermediate calculation process |
| Reward generation | 7.25E-06 | Generates tracking reward as per Equation (17) |
| Next state generation | 1.80E-05 | Generates next state |
| New-old alternation 2 | 1.61E-06 | Intermediate calculation process |

**Table 9 Transmission of Memory Buffer – Task Duration and Function**

| Task | Typical Duration (s) | Function |
| --- | --- | --- |
| Data packing | 5.47E-05 | Generates binary data packets |
| Data transmission | 5.71E-05 | Data transmission via network interface |

**Table 10 Transmission of Action – Task Duration and Function**

| Task | Typical Duration (s) | Function |
| --- | --- | --- |
| New-old alternation 3 | 3.50E-06 | Intermediate calculation process |
| Action packing and transmission to the cloud | 4.43E-05 | Data transmission via network interface |

Based on the above observations:

1. Impact of Reinforcement Learning Control on I/O Processes: The reinforcement learning control process primarily affects the I/O components, posing significant challenges to real-time performance. The Sensor data unpacking process, Data packing, Data transmission, Transmission of Memory Buffer, and Transmission of Action account for up to 68.9% of the total execution time, making real-time implementation highly challenging.

2. Benefits of Framework Optimization: Utilizing the conventional CRL2E algorithm[21], the unoptimized execution time was 8.43E-04 seconds. After optimizing the framework as described in Section 4.1, the execution time was reduced by approximately half, ranging from 3.29E-04 to 5.14E-04 seconds.

3. Advantages of Direct Matrix Operations Over PyTorch: Implementing the edge inference framework defined in Section 4.6 significantly enhances algorithm performance by enabling direct matrix operations without the need for tensorization and de-tensorization. A notable improvement is the reduction of the Reception, Analysis, and Loading of Weight time by three orders of magnitude. Additionally, the State tensorization and Action de-tensorization processes experienced a two to three orders of magnitude decrease in processing time. Consequently, the average CPU time was improved by more than three times compared to existing studies[49].

## D. Question 2: How is the applicability of the CRL framework to various classical controllers evaluated under real hardware interference conditions?

To validate the performance of the proposed CRL2RT algorithm under real hardware interference conditions, experiments were conducted based on the operating conditions defined in Section 5.3. The results are illustrated in Fig 10 to Fig 12, which includes the following configurations:

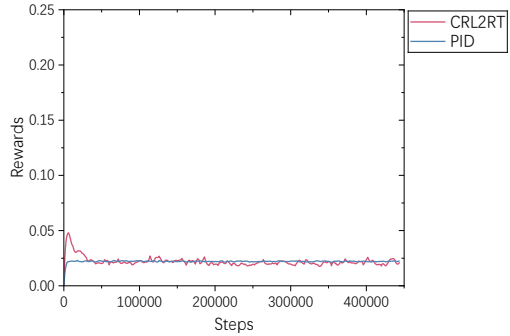

(a) Load 1, 20 Hz configuration

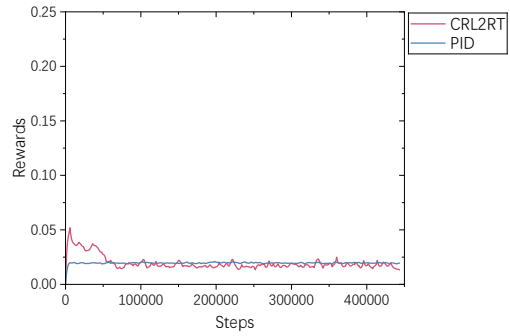

(b) Load 2, 20 Hz configuration

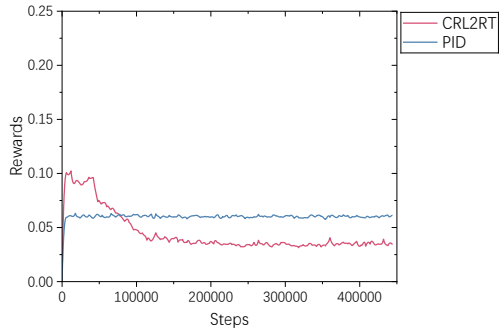

(c) Load 1, 40 Hz configuration

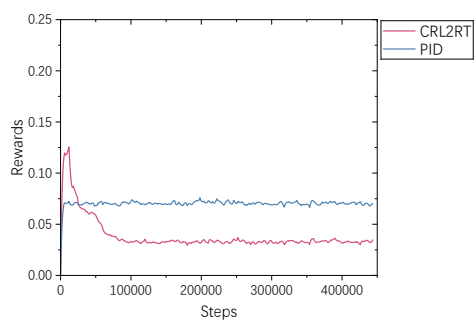

(d) Load 2, 40 Hz configuration

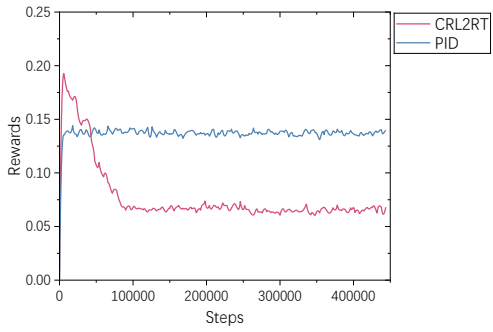

(e) Load 1, 60 Hz configuration

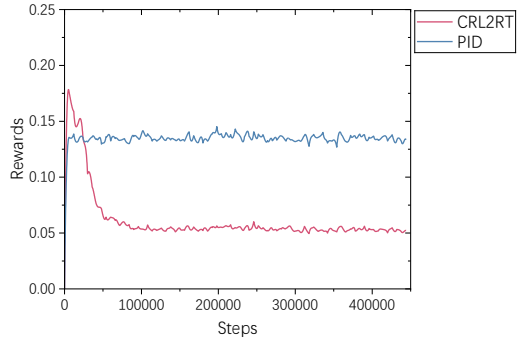

(f) Load 2, 60 Hz configuration

**Fig. 10. Algorithm Performance under PID Controller Conditions**

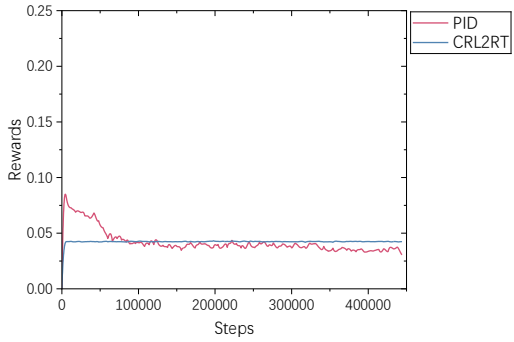

(a) Load 1, 20 Hz configuration

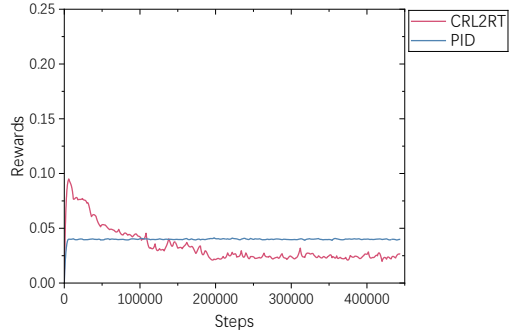

(b) Load 2, 20 Hz configuration

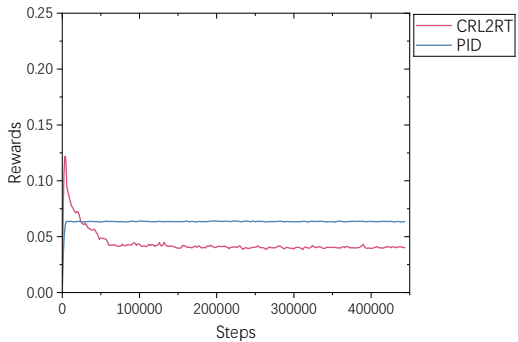

(c) Load 1, 40 Hz configuration

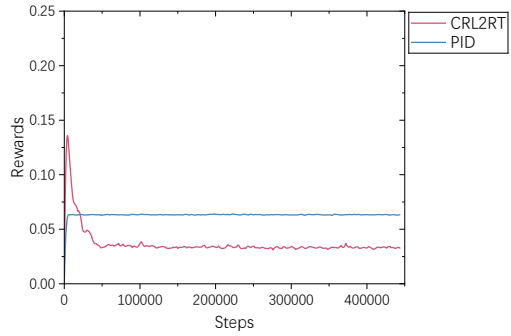

(d) Load 2, 40 Hz configuration

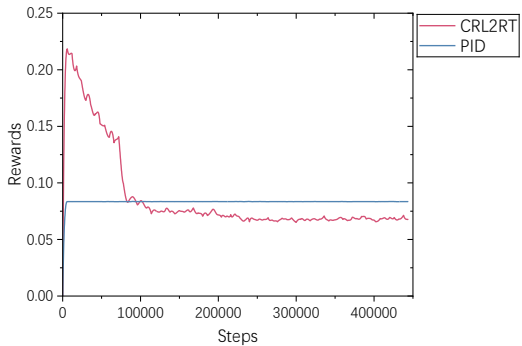

(e) Load 1, 60 Hz configuration

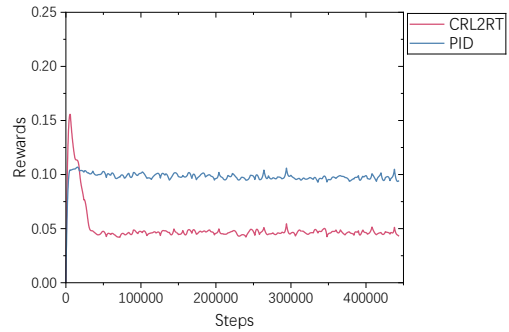

(f) Load 2, 60 Hz configuration

**Fig. 11. Algorithm Performance under Adaptive PID Controller Conditions**

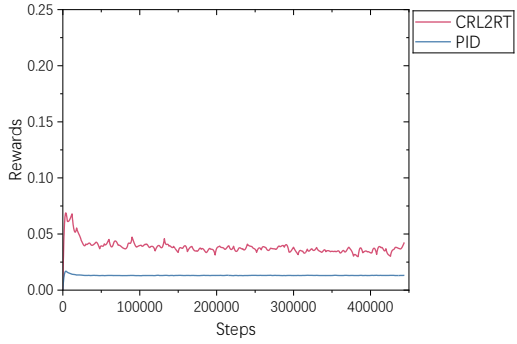
(a) Load 1, 20 Hz configuration

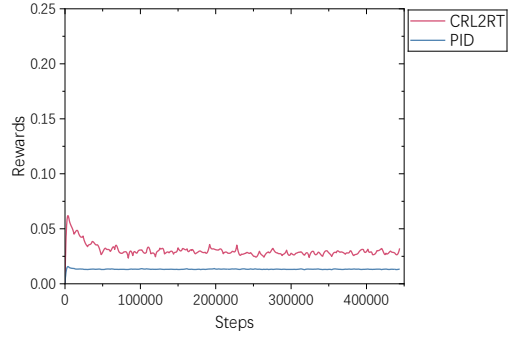
(b) Load 2, 20 Hz configuration

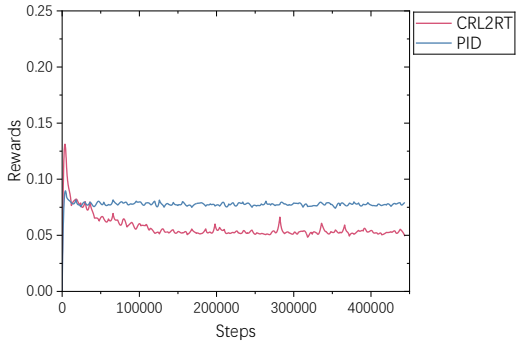
(c) Load 1, 40 Hz configuration

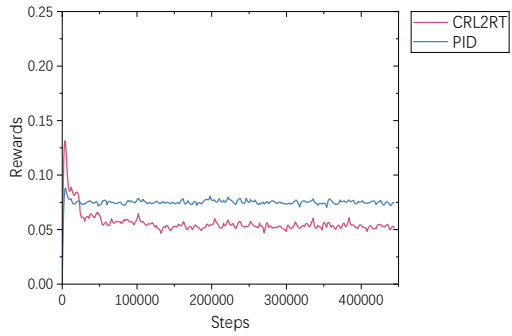
(d) Load 2, 40 Hz configuration

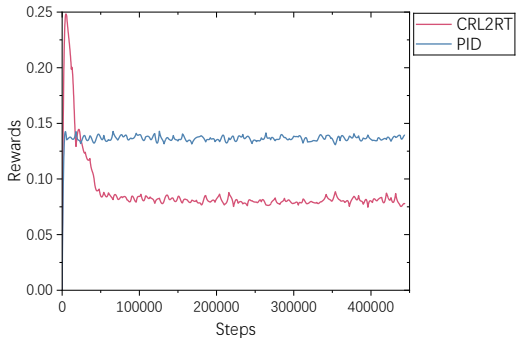
(e) Load 1, 60 Hz configuration

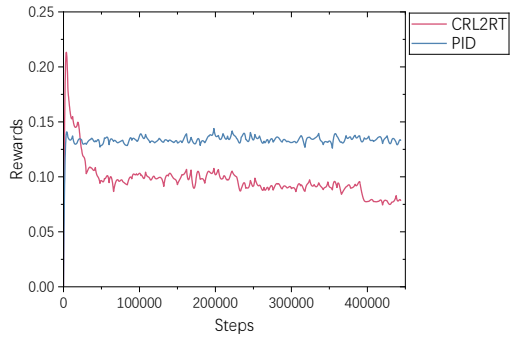
(f) Load 2, 60 Hz configuration

**Fig. 12. Algorithm Performance under MRAC Controller Conditions**

**Table 11 End Performance of the Last Quarter of the Training Process**

| Classical Controller Type | Condition | Traditional Control | CRL Mode | Tracking Error Level (%) |
|---|---|---|---|---|
| PID Controller | Load 1, 20 Hz configuration | 0.022087 | 0.020808 | 5.8 |
| | Load 1, 40 Hz configuration | 0.060237 | 0.034895 | 42.1 |
| | Load 1, 60 Hz configuration | 0.137391 | 0.06548 | 52.3 |
| | Load 2, 20 Hz configuration | 0.019569 | 0.017646 | 9.8 |
| | Load 2, 40 Hz configuration | 0.070401 | 0.033128 | 52.9 |
| | Load 2, 60 Hz configuration | 0.134749 | 0.052894 | 60.7 |
| Adaptive PID Controller | Load 1, 20 Hz configuration | 0.042419 | 0.034899 | 17.7 |
| | Load 1, 40 Hz configuration | 0.063518 | 0.040488 | 36.3 |
| | Load 1, 60 Hz configuration | 0.083435 | 0.068146 | 18.3 |
| | Load 2, 20 Hz configuration | 0.039932 | 0.02389 | 40.2 |
| | Load 2, 40 Hz configuration | 0.063286 | 0.033169 | 47.6 |
| | Load 2, 60 Hz configuration | 0.096686 | 0.046247 | 52.2 |
| | Load 1, 20 Hz configuration | 0.01303 | 0.035438 | -172.0 |
| MRAC Controller | Load 1, 40 Hz configuration | 0.077608 | 0.052866 | 31.9 |
| | Load 1, 60 Hz configuration | 0.136805 | 0.080656 | 41.0 |
| | Load 2, 20 Hz configuration | 0.013174 | 0.02836 | -115.3 |
| | Load 2, 40 Hz configuration | 0.074944 | 0.053581 | 28.5 |
| | Load 2, 60 Hz configuration | 0.134046 | 0.085128 | 36.5 |

Based on the above figures and table 11, the following observations can be made:

1. The proposed CRL2RT algorithm demonstrated rapid convergence across all three classical controller conditions. The convergence trends closely matched the Finite-Time Lyapunov Convergence conditions established in Section 4.3.

2. Algorithms such as MRAC, due to their adaptive nature, exhibited notable performance enhancements under both Load 1 and Load 2 configurations at 40 Hz and 60 Hz, the CRL2RT algorithm combined with all three classical controllers achieved tracking performance improvements ranging from 18.3% to 60.7%, highlighting the broad applicability of the CRL2RT algorithm.

3. At the 20 Hz configuration, the CRL2RT algorithm showed relative performance enhancements when combined with PID and Adaptive PID controllers, further demonstrating its versatility.

4. However, when combined with the MRAC controller under the 20 Hz configuration, the CRL2RT algorithm exhibited convergence trends but did not significantly outperform the MRAC controller. This is attributed to the adaptive capabilities of the MRAC algorithm, which provided better performance without offering diverse data, thereby affecting the $F_{explore}(t)$ and subsequently the convergence process of the CRL2RT algorithm.

The theoretical reasons for the observed phenomena are as follows:

1. From the Perspective of the Time-Interleaved Module: Besides the divergence suppression and exploration limitation analyzed in Section 5.4, the replay buffer recording mode defined in equation (45) can be regarded as imitation learning without covariate shift[88]. The traditional control module within the Time-Interleaved module provides high-quality example data before the RL algorithm surpasses the performance of classical control methods, supporting policy updates effectively.

2. From the Rule-Based Policy Composer Strategy Perspective: The theoretical basis for enhanced convergence speed lies in the Finite-Time Lyapunov Convergence Conditions demonstrated in Section 4.3.2. By prioritizing convergence speed in its restructuring, the Policy Composer significantly improves the algorithm's ability to converge rapidly.

## VII. Conclusion

This paper addresses the challenges of real-time control for a direct-drive tandem-wing experimental platform using a novel reinforcement learning framework, CRL2RT. The study's primary contributions are summarized as follows:

1. Proposed CRL2RT Algorithm for Real-Time Reinforcement Learning: The CRL2RT algorithm enhances the Concerto Reinforcement Learning framework by optimizing weight updates and inference strategies for real-time applications. It achieves over 2500 Hz control frequency on general-purpose CPUs, making it the world's fastest reinforcement learning algorithm with online weight updates under such conditions.

2. Integration with Traditional Controllers: The CRL2RT algorithm demonstrated wide applicability by integrating with three types of classical controllers (PID, Adaptive PID, and MRAC). This integration ensured safety and improved convergence rates through the Time-Interleaved module, satisfying Lipschitz continuity and finite-time Lyapunov stability conditions.

3. Experimental Performance Validation: Extensive experiments across six operating conditions (varying frequencies and loads) validated the algorithm's effectiveness. The CRL2RT algorithm consistently enhanced tracking performance, achieving 18.3%–60.7% improvement under challenging dynamic conditions.

4. Theoretical Insights into Convergence: The Rule-Based Policy Composer improved convergence speed by leveraging finite-time Lyapunov stability conditions and exploiting gradient-domain Laplace transforms. The findings demonstrated the algorithm's robustness and adaptability under multiple random operating conditions.

5. Deployment and Real-Time Performance: The study highlighted the critical role of optimizing the inference process. By transitioning to direct matrix operations and redesigning the inference framework, the algorithm achieved a threefold improvement in inference speed compared to conventional implementations.

Despite the promising results, certain limitations were observed, particularly in scenarios involving adaptive controllers like MRAC, where CRL2RT's performance did not significantly surpass the adaptive capabilities of MRAC alone. Future work will focus on further refining the integration mechanisms and exploring adaptive strategies to enhance the algorithm's performance in conjunction with various adaptive control systems.

Future research should concentrate on applying the CRL2RT algorithm to flight control scenarios of dragonfly-inspired and other biomimetic aerial vehicles and exploring the universality of the CRL2RT algorithm when integrated with additional existing traditional control algorithms. The CRL2RT algorithm holds significant potential for application across various domains. In mobile robotics, it could be extended to trajectory tracking control for platforms such as quadruped robots and humanoid robots. Furthermore, it shows promise for precision motion control in fields like precision manufacturing.

Overall, the CRL2RT algorithm represents a significant advancement in real-time control for biomimetic aerial vehicles, offering a robust solution to the challenges posed by nonlinear, unsteady aerodynamic interactions in tandem-wing configurations.

## Acknowledgements

Data will be made available on reasonable request.

This work is financially supported by the Shenzhen Science and Technology Program and Research under Grant No. JCYJ20220530161808018, and Guangdong Basic and Applied Basic Research Foundation under Grant No. 2023A1515010776. Key R&D Program in Shaanxi Province of China under Grant No. 2023-YBGY-372. "The Fundamental Research Funds for the Central Universities"

## Conflict of interests

The authors declare that they have no conflict of interest.

60. Kerbel, L., Ayalew, B., and Ivanco, A. "Adaptive policy learning for data-driven powertrain control with eco-driving," *Engineering Applications of Artificial Intelligence* Vol. 124, 2023, p. 106489.
doi: https://doi.org/10.1016/j.engappai.2023.106489
61. Tsybina, Y. A., Gordleeva, S. Y., Zharinov, A., Kastalskiy, I., Ermolaeva, A., Hramov, A., and Kazantsev, V. "Toward biomorphic robotics: A review on swimming central pattern generators," *Chaos, Solitons & Fractals* Vol. 165, 2022, p. 112864.
doi: https://doi.org/10.1016/j.chaos.2022.112864
62. ZHANG, M., Song, B., Yang, X., Wang, L., and Lang, X. "Development of Minimal Biorobotic Stealth Distance and its Application in the Design of Direct-Drive Dragonfly-Inspired Aircraft," *Available at SSRN 4930951*.
doi: https://doi.org/10.48550/arXiv.2410.15559
63. Potosakis, N., Paraskevopoulos, E., and Natsiavas, S. "Application of an augmented Lagrangian approach to multibody systems with equality motion constraints," *Nonlinear Dynamics* Vol. 99, No. 1, 2020, pp. 753-776.
doi: https://doi.org/10.1007/s11071-019-05059-6
64. Dong, S., Otsuka, K., and Makihara, K. "Hamiltonian formulation with reduced variables for flexible multibody systems under linear constraints: Theory and experiment," *Journal of Sound and Vibration* Vol. 547, 2023, p. 117535.
doi: https://doi.org/10.1016/j.jsv.2022.117535
65. Kumar, A., Li, Z., Zeng, J., Pathak, D., Sreenath, K., and Malik, J. "Adapting Rapid Motor Adaptation for Bipedal Robots," *2022 IEEE/RSJ International Conference on Intelligent Robots and Systems (IROS)*. 2022, pp. 1161-1168.
66. Caetano, J., Weehuizen, M., De Visser, C., De Croon, G., and Mulder, M. "Rigid-body kinematics versus flapping kinematics of a flapping wing micro air vehicle," *Journal of Guidance, Control, and Dynamics* Vol. 38, No. 12, 2015, pp. 2257-2269.
doi: https://doi.org/10.2514/1.G000923
67. Garg, K., Arabi, E., and Panagou, D. "Prescribed-time convergence with input constraints: A control Lyapunov function based approach," *2020 American Control Conference (ACC)*. IEEE, 2020, pp. 962-967.
68. Silver, D., Lever, G., Heess, N., Degris, T., Wierstra, D., and Riedmiller, M. "Deterministic policy gradient algorithms," *International conference on machine learning*. Pmlr, 2014, pp. 387-395.
69. Sutton, R. S., and Barto, A. G. *Reinforcement learning: An introduction*: MIT press, 2018.
70. Dalal, G., Szorenyi, B., and Thoppe, G. "A tale of two-timescale reinforcement learning with the tightest finite-time bound," *Proceedings of the AAAI Conference on Artificial Intelligence*. Vol. 34, 2020, pp. 3701-3708.
71. Xu, T., and Liang, Y. "Sample complexity bounds for two timescale value-based reinforcement learning algorithms," *International Conference on Artificial Intelligence and Statistics*. PMLR, 2021, pp. 811-819.
72. Momcilovic, S., Ilic, A., Roma, N., and Sousa, L. "Dynamic load balancing for real-time video encoding on heterogeneous CPU+ GPU systems," *IEEE Transactions on Multimedia* Vol. 16, No. 1, 2013, pp. 108-121.
doi: 10.1109/TMM.2013.2284892
73. Luo, S., Deng, F., and Yu, X. "Unified stability analysis for Itô stochastic systems: From almost surely asymptotic to finite-time convergence," *IEEE Transactions on Automatic Control* Vol. 67, No. 1, 2021, pp. 406-412.
doi: 10.1109/TAC.2021.3057990
74. Meunier, D., Lambiotte, R., and Bullmore, E. T. "Modular and hierarchically modular organization of brain networks," *Frontiers in neuroscience* Vol. 4, 2010, p. 7572.
doi: https://doi.org/10.3389/fnins.2010.00200
75. Léger, C., Hamon, G., Nisioti, E., Hinaut, X., and Moulin-Frier, C. "Evolving Reservoirs for Meta Reinforcement Learning," *International Conference on the Applications of Evolutionary Computation (Part of EvoStar)*. Springer, 2024, pp. 36-60.
76. Jang, K., Bang, H., and Kim, Y. "Mitigating Time-Delay in Nonlinear Dynamics Inversion for Multirotor Unmanned Aerial Vehicles," *Journal of Guidance, Control, and Dynamics*, 2024, pp. 1-15.
doi: https://doi.org/10.2514/1.G007567
77. Wang, H., Wu, J., Chen, Z., He, R., and Li, H. "Vehicle yaw stability model predictive control strategy for dynamic and multi-objective requirements." SAE Technical Paper, 2024.
78. Barbara, N. H., Wang, R., and Manchester, I. R. "On Robust Reinforcement Learning with Lipschitz-Bounded Policy Networks," *arXiv preprint arXiv:2405.11432*, 2024.